\documentclass{article} 
\usepackage{iclr2027_conference,times}

\usepackage[utf8]{inputenc}
\usepackage[T1]{fontenc}

\usepackage{booktabs}
\usepackage{multirow}
\usepackage{wrapfig}
\usepackage{amsfonts}
\usepackage{amsmath}
\usepackage{amsthm}
\usepackage{graphicx}
\usepackage{algorithm}
\usepackage{algorithmic}
\usepackage{placeins}
\usepackage{microtype}
\usepackage{xcolor}
\usepackage{nicefrac}

\usepackage{hyperref}
\hypersetup{
  hypertexnames=false,
  colorlinks=true,
  linkcolor=blue!60!black,
  citecolor=blue!60!black,
  urlcolor=blue!60!black
}
\usepackage{url}

\newtheorem{assumption}{Assumption}[section]
\newtheorem{lemma}{Lemma}[section]
\newtheorem{proposition}{Proposition}[section]
\newtheorem{corollary}{Corollary}[section]
\newtheorem{remark}{Remark}[section]

\title{BGM-IV: AI-Powered Bayesian Generative Modeling for Instrumental Variable Regression with High-Dimensional Covariates}

\author{Guyue Luo$^{1}$, Haidong Lu$^{2,3,4}$, Andrew J. Loza$^{2,5}$, Qiao Liu$^{1}$\thanks{Corresponding author: qiao.liu@yale.edu} \\
  $^{1}$Department of Biostatistics, Yale University \\
  $^{2}$Department of Biomedical Informatics and Data Science, Yale University \\
  $^{3}$Department of Internal Medicine, Yale University \\
  $^{4}$Department of Chronic Disease Epidemiology, Yale University \\
  $^{5}$Department of Pediatrics, Yale University \\
  New Haven, CT 06510, USA \\
  \texttt{\{guyue.luo,haidong.lu,andrew.loza,qiao.liu\}@yale.edu}}

\iclrfinalcopy 
\begin{document}

\maketitle
\lhead{Preprint}

\begin{abstract}
Instrumental-variable (IV) regression enables causal estimation under endogeneity, but modern IV problems often involve nonlinear structural effects and high-dimensional covariates. Existing methods typically operate in observed or generic learned feature spaces, and they often yield point estimates without uncertainty quantification. We introduce BGM-IV, a Bayesian generative modeling approach that performs nonlinear IV regression through posterior inference in a causally structured latent space. BGM-IV separates covariate variation by the role in the treatment and outcome mechanism, and accounts for endogeneity through an IV-integrated pseudo-likelihood that averages over instrument-induced treatment variation. The resulting model provides both structural-function estimates and predictive intervals for outcomes under intervention. Across various benchmark datasets, BGM-IV outperforms existing nonlinear IV methods overall, with significant gains in high-dimensional settings, while achieving near-nominal predictive coverage. These results highlight structured latent generative modeling as a flexible approach to uncertainty-aware IV inference with rich covariates. The code of BGM-IV is available at \url{https://github.com/liuq-lab/BGM-IV}.
\end{abstract}

\section{Introduction}
Estimating causal effects from observational data is challenging when treatment is endogenous. Unobserved confounders jointly affect treatment and outcome and a predictive model may fit the observed outcome well but fail to recover the effect of an intervention~\citep{imbens2015causal,hernan2025causal,Hernán02012019}. Instrumental variables (IVs) address this problem by exploiting exogenous variation that shifts treatment while affecting the outcome only through treatment~\citep{angrist1996iv,angrist2009mostly,angrist1995identification}. IV estimation is therefore inherently indirect, as the structural effect must be recovered from instrument-induced treatment variation.

Modern IV problems are further complicated by nonlinear structural effects and high-dimensional covariates, such as images, electronic health records, or molecular measurements. Rich covariates may contain useful information about confounding, but treating them directly is statistically and computationally difficult~\citep{chernozhukov2022automatic}. Simply plugging predictive models into an IV estimator is often not enough as poor handling of high-dimensional covariates can lead to biased causal estimates~\citep{damour2021overlap,chernozhukov2018dml}. Addressing such settings requires both flexible nonlinear causal effect modeling and a representation that separates causally relevant covariate information from nuisance variation~\citep{johansson2016learning,louizos2017causal,shalit2017estimating,scholkopf2021toward}.

Existing work has largely expanded the scope of IV estimation beyond classical linear models~\citep{wu2025instrumental}. Two-stage least squares (2SLS) remains widely used for its simplicity and interpretability under linear specifications~\citep{angrist1996iv,stock2003ivhistory}. Nonparametric IV methods relax these restrictions through series, sieve, and regularized inverse formulations~\citep{newey2003nonparametric,blundell2007semi,darolles2011nonparametric,chen2012conditional}. Recent machine learning IV methods further increase flexibility: KIV generalizes 2SLS to reproducing kernel Hilbert spaces~\citep{singh2019kiv}, DeepIV integrates the outcome mean over an estimated conditional treatment distribution~\citep{hartford2017deepiv}, DFIV learns deep treatment and instrument features within 2SLS by alternating between the two stages~\citep{xu2021dfiv}, DeepGMM and related estimators extend the generalized method of moments to conditional moment restrictions through minimax optimization~\citep{hansen1982gmm,bennett2019deepgmm,dikkala2020minimax}, DML-IV applies double machine learning to mitigate regularization bias of the two-stage estimators~\citep{shao2024dmliv}, and FIL combines kernelized IV regression with adaptive instrument learning~\citep{wang2022fil}.

Despite this progress, nonlinear IV estimation remains challenging with high-dimensional covariates, where causally relevant information may be dominated by nuisance variation. Existing approaches typically operate on observed covariates or generic learned features, without explicitly capturing the roles of the covariates in the treatment and outcome mechanisms. In addition, most methods emphasize point estimation, while uncertainty quantification for outcomes under intervention remains limited. These gaps motivate an AI-powered generative framework that separates shared, treatment-specific, outcome-specific, and nuisance variation while supporting uncertainty-aware IV inference.

In this work, we propose BGM-IV, a Bayesian generative modeling approach for nonlinear IV regression. The central idea is to perform IV regression through posterior inference in a causally structured latent space, rather than directly in observed or generic learned feature space. By partitioning the structured latent space into independent subspaces that play different roles in the causal modeling, BGM-IV provides a principled way for estimating causal effects with IVs with uncertainty quantification. Our main contributions are summarized as follows.

\begin{itemize}
\item BGM-IV bridges the flexibility of deep generative models with principled posterior inference for nonlinear IV regression. The structured latent space partitions high-dimensional covariate variation based on the causal roles, and a novel IV-integrated pseudo-likelihood accounts for endogeneity by integrating the outcome density over the IV-induced treatment distribution.

\item BGM-IV provides uncertainty quantification together with the point estimate. Posterior inference over the covariate-conditioned latent state yields structural-function estimates and predictive intervals for outcomes under intervention.

\item Across diverse benchmarks, BGM-IV achieves the best performance compared to strong nonlinear IV baselines in nearly all evaluated settings, with the largest gains under high-dimensional covariates, while its predictive intervals attain near-nominal coverage.
\end{itemize}

\section{Problem Setup}
\label{sec:problem_setup}

We observe $i.i.d.$\ samples $\{(X^{(i)},Y^{(i)},V^{(i)},W^{(i)})\}_{i=1}^n$ from an observational distribution, where $X\in\mathcal{X}$ is the treatment (or exposure), $Y\in\mathbb{R}$ is the outcome, $V\in\mathcal{V}$ denotes the high-dimensional covariates, and $W\in\mathcal{W}$ is an instrumental variable. Our target is the structural function
\[
g_0(x,\mathbf{v})=\mathbb{E}\!\left[Y \mid \mathrm{do}(X=x),\, V=\mathbf{v}\right],
\]
using Pearl's intervention notation~\citep{pearl2009causality}. Under the structural formulation, we write
\[
Y = g_0(X,V) + \epsilon.
\]
The challenge is that treatment is endogenous: in general,
$\mathbb{E}[\epsilon \mid X,V]\neq 0$,
so the observational regression function $\mathbb{E}[Y\mid X=x,V=\mathbf{v}]$ generally differs from the causal target $g_0(x,\mathbf{v})$.

Conditional on $V$, the instrument satisfies three standard conditions. Relevance requires the conditional distribution $P(X\mid V,W)$ to vary with  $W$~\citep{bound1995problems}. Exclusion requires the instrument to affect the outcome only through its effect on treatment. Exogeneity requires the instrument to be mean-independent of the structural error conditional on the observed covariates (i.e., conditional mean exogeneity), $\mathbb{E}[\epsilon\mid V,W]=0$.\footnote{This is a sufficient conditional mean exogeneity restriction and does not require
$W\perp\epsilon\mid V$.} Under these conditions,
\[
\mathbb{E}[Y\mid V,W]
=\mathbb{E}[g_0(X,V)\mid V,W]
=\int g_0(x,V)\,dP(x\mid V,W).
\]
This moment equation shows that the instrument does not identify the structural function $g_0(x,\mathbf{v})$ directly. Instead, it identifies how $g_0$ averages over the treatment distribution induced by the instrument, conditional on the covariates. In nonparametric IV, recovering $g_0$ requires conditions such as completeness of the conditional expectation operator~\citep{newey2003nonparametric,hall2005nonparametric,carrasco2007linear,d2011completeness}.

The remaining challenge is how to model the causal target $g_0(x,\mathbf{v})$ and the observable moment equation induced by the IV assumptions, particularly when the covariates $V$ are high-dimensional. The next section addresses this by introducing the latent generative structure used in BGM-IV.

\section{Method}
\label{sec:method}

\subsection{Causally Structured Bayesian Generative Model}

Inspired by recent advances in structural latent space modeling~\citep{liu2026bgm,liu2026causalbgm,liu2026missingness}, we further extend it to nonlinear IV regression. Specifically, we partition the latent space $Z=(Z_0,Z_1,Z_2,Z_3)$ to account for distinct causal roles. The shared block $Z_0$ enters both the treatment and outcome mechanisms and captures the observed confounding structure. 
$Z_1$ enters only the outcome mechanism, $Z_2$ enters only the treatment mechanism, and $Z_3$ is a nuisance block that explains the remaining covariate variation.

In the inference stage, the structured latent state is inferred from the observed covariates $V$ alone. BGM-IV addresses the unobserved confounding between $X$ and $Y$ through the instrument and the
IV-integrated objective. This partition provides a causally structured representation of how high-dimensional covariates impact treatment and outcome. 

\begin{figure}[t]
  \centering
  \includegraphics[width=1\linewidth]{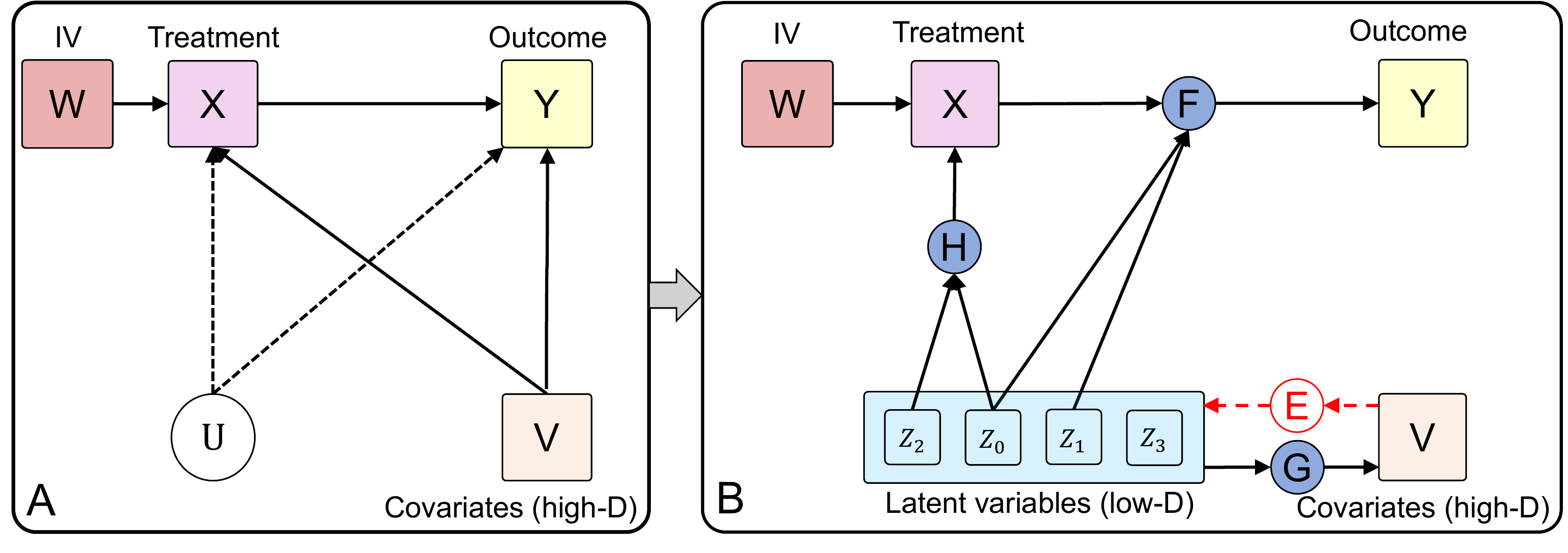}
  \caption{Illustration of the BGM-IV framework. (A) Causal graph for the IV setting. $U$ is the unobserved confounder. (B) The causal graph in BGM-IV model. Rectangles denote variables and circles denote learned neural networks. $G$, $H$, and $F$ denote the covariate, treatment, and outcome generative models, respectively. The encoder $E$ is used in the EGM initialization and is not updated afterwards.}
  \label{fig:bgm_dag}
  \vspace{-10pt}
\end{figure}

The generative
components of BGM-IV (Figure~\ref{fig:bgm_dag}) are specified as
\begin{equation}
Z\sim p_Z,\qquad
V\sim p_\theta(\mathbf{v}\mid \mathbf{z}),\qquad
X\sim p_\phi(x\mid w,\mathbf{z}_0,\mathbf{z}_2),\qquad
Y\sim p_\omega(y\mid x,\mathbf{z}_0,\mathbf{z}_1).
\label{eq:bgm_iv_joint_model}
\end{equation}
Here $p_Z$ is the latent prior, which is set to be $\mathcal{N}(0,I)$ by default. $p_\theta$, $p_\phi$, and $p_\omega$ are neural generative
models for the covariates, treatment, and structural outcome,
respectively. The instrumental variable $W$ enters only through the treatment mechanism, reflecting the exclusion restriction. Because $X$ is endogenous,
$p_\omega$ is not fitted as the observational conditional distribution
of $Y$ given $X$. Instead, its structural mean is learned through the
IV-integrated pseudo-likelihood introduced in
Section~\ref{sec:pseudo_likelihood}.


By default, the three generative models $p_\theta$, $p_\phi$, and $p_\omega$ are specified as multivariate normal distributions for continuous variables and Bernoulli or Categorical distributions for discrete variables. In a typical continuous setting, we parameterize the models as

\begin{equation}
\left\{
\begin{aligned}
V \mid Z &\sim \mathcal{N}\!\big(\boldsymbol{\mu}_\theta(\mathbf{z}),\, \boldsymbol{\Sigma}_\theta(\mathbf{z})\big),\\
X \mid W, Z_0, Z_2
&\sim \mathcal{N}\!\big(\mu_\phi(w,\mathbf{z}_0,\mathbf{z}_2),\, \sigma_\phi^2(w,\mathbf{z}_0,\mathbf{z}_2)\big),\\
Y \mid X, Z_0, Z_1
&\sim \mathcal{N}\!\big(\mu_\omega(x,\mathbf{z}_0,\mathbf{z}_1),\, \sigma_\omega^2(x,\mathbf{z}_0,\mathbf{z}_1)\big).
\end{aligned}
\right.
\label{eq:continuous_model}
\end{equation}

Here ($\boldsymbol{\mu}_\theta$, $\boldsymbol{\Sigma}_\theta$), ($\mu_\phi$, $\sigma_\phi^2$) and ($\mu_\omega$, $\sigma_\omega^2$) denote learnable functions, 
represented by three neural networks ($G$, $H$, and $F$) each with two output heads. For further simplification,
we use a diagonal covariance structure for the covariate model $\boldsymbol{\Sigma}_\theta(\mathbf{z})=\mathrm{diag}\!\big(\sigma_{\theta,1}^2(\mathbf{z}),\dots,\sigma_{\theta,p}^2(\mathbf{z})\big)$.
For binary treatments, the treatment model is replaced by a Bernoulli distribution
$X \mid W, Z_0, Z_2
\sim
\mathrm{Bernoulli}\!\big(\pi_\phi(w,\mathbf{z}_0,\mathbf{z}_2)\big)$
where $\pi_\phi(w,\mathbf{z}_0,\mathbf{z}_2)=\mathrm{logit}^{-1}(H(w,\mathbf{z}_0,\mathbf{z}_2; \phi))$ represents the propensity score.
Analogous Bernoulli or categorical parameterizations can be used for discrete outcomes or covariates.

\subsection{IV-Integrated Pseudo-Likelihood}
\label{sec:pseudo_likelihood}

With endogenous treatment, directly training the outcome generator through $p_\omega(y\mid x,\mathbf{z}_0,\mathbf{z}_1)$ using the observed treatment can bias estimation of the structural function $g_0$ as the observed treatment reflects both causal variation and unobserved confounding. BGM-IV therefore uses an IV-integrated working likelihood
for outcome learning. Specifically, it integrates the outcome density over the treatment distribution induced by the instrument, which is denoted as:
\begin{equation}
p_{\mathrm{IV}}(y\mid w,\mathbf{z}_0,\mathbf{z}_1,\mathbf{z}_2)
=
\int p_\omega(y\mid x,\mathbf{z}_0,\mathbf{z}_1)\,
p_\phi(x\mid w,\mathbf{z}_0,\mathbf{z}_2)\,dx .
\label{eq:iv_integrated_likelihood}
\end{equation}
This is our derived IV-integrated pseudo-likelihood term, which is also related to the integrated treatment-distribution used by DeepIV~\citep{hartford2017deepiv}, but the objectives differ in both form and role. DeepIV minimizes a squared loss between the observed outcome and the response averaged over the estimated first-stage treatment distribution. BGM-IV instead integrates the outcome density
first and uses the resulting mixture density as a component of latent quasi-posterior inference. Consequently, the IV-integrated term informs both the structural outcome generator and the update of the latent states during training. 

Let $\mu_\omega$ denote the mean of $p_\omega$, the conditional mean implied by Eq.~\eqref{eq:iv_integrated_likelihood} is
\begin{equation}
\mathbb{E}_{\omega,\phi}[Y\mid W=w,Z=\mathbf z]
=
\int \mu_\omega(x,\mathbf z_0,\mathbf z_1)
 p_\phi(x\mid w,\mathbf z_0,\mathbf z_2)\,dx,
\label{eq:iv_conditional_mean}
\end{equation}
which has the same conditional-expectation operator form as the IV moment in Section~\ref{sec:problem_setup}.

\subsection{Identification and Population Target}
\label{sec:identification}

Appendix~\ref{app:iv_population_justification} analyzes the population
structural-mean target of the conditional moment equation
\eqref{eq:iv_conditional_mean} for a covariate-measurable
latent representation
$A=(Z_0,Z_1,Z_2)=\varphi(V)$, with
$A_Y=(Z_0,Z_1)$ and $A_X=(Z_0,Z_2)$, such that
$g_0(x,V)=m_0(x,A_Y)$.

\textbf{Mean identification.}
Under latent conditional mean exogeneity,
$\mathbb{E}[\epsilon\mid W,A]=0$, the structural mean satisfies the latent nonparametric IV moment equation with the same conditional-expectation operator form as Eq.~\eqref{eq:iv_conditional_mean}. If the corresponding
true conditional expectation operator is complete on the
considered function class, the structural mean $m_0$ is the unique
solution of this moment equation on the identified treatment support
(Proposition~\ref{prop:latent_iv_moment}).

\textbf{Consistency of structural prediction.}
Proposition~\ref{prop:final_map_selection} bounds the IV-moment
risk of the selected covariate-only MAP predictor. Under final-candidate approximation, first-stage, empirical, and Monte Carlo error control, and controlled ill-posedness on the candidate class~\citep{xu2021dfiv,dikkala2020minimax}, Corollary~\ref{cor:quasi_posterior_consistency} translates this risk bound into $L^2(P_{X,V})$ consistency of the selected structural-mean predictor.

\subsection{Tempered Latent Quasi-Posterior}
\label{sec:quasi_posterior}

We combine the IV-integrated term with the latent prior, the covariate likelihood, and the first-stage treatment likelihood to define the quasi-posterior as
\begin{equation}
q_{\mathrm{IV}}(\mathbf{z}\mid x,y,\mathbf{v},w)
\propto
p(\mathbf{z})\,p_\theta(\mathbf{v}\mid \mathbf{z})\,
p_\phi(x\mid w,\mathbf{z}_0,\mathbf{z}_2)\,
p_{\mathrm{IV}}(y\mid w,\mathbf{z}_0,\mathbf{z}_1,\mathbf{z}_2)^{\gamma},
\qquad \gamma\in[0,1].
\label{eq:iv_quasi_posterior}
\end{equation}
$\gamma$ is a tempering parameter used to control how strongly the IV-integrated outcome term
influences latent-state inference~\citep{https://doi.org/10.1111/rssb.12158}. We set $\gamma=0.01$ by default and report a sensitivity analysis over $\gamma$ in Appendix~\ref{app:gamma_sensitivity}.

During training, Eq.~\eqref{eq:iv_quasi_posterior} is used to update subject-specific latent variables using $Y$, raising the concern that they could absorb the structural error and reintroduce endogeneity. The key distinction is that these outcome-informed latent states are not reused for structural prediction. In the inference stage, BGM-IV re-infers the latent state from $V$ alone through the covariate-only posterior in
Section~\ref{sec:model_inference}. Thus, for a new subject, the latent state entering the final structural predictor preserves
conditional mean exogeneity
(Lemma~\ref{lem:latent_exogeneity}). Appendix~\ref{app:latent_exogeneity} empirically checks that the resulting evaluation states show no detectable association with the structural error or the unobserved confounder.

\subsection{Stochastic Optimization Training}
\label{sec:training}

We first develop an IV-aware initialization strategy based on EGM~\citep{liu2024encoding}, which helps initialize both the model parameters and latent states (see details in Appendix~\ref{app:detailed_algorithm}). 
Since the joint posterior distribution of latent variables and model parameters is not tractable under BGM-IV, we designed an iterative stochastic optimization procedure that alternates between updating subject-specific latent variables and updating the generator parameters. At each iteration, we draw a mini-batch $\mathcal{B}$. Each iteration consists of four steps:

(a) Given the current latent states $\{\mathbf{z}^{(i)}|i=1,...,n\}$, the covariate generative model is updated using
\begin{equation}
\mathcal{L}_V(\theta)
=
-\frac{1}{|\mathcal{B}|}
\sum_{i\in\mathcal{B}}
\log p_\theta(\mathbf{v}^{(i)} \mid \mathbf{z}^{(i)}).
\label{eq:loss_v_batch}
\end{equation}

(b) Given the current latent states $\{\mathbf{z}^{(i)}|i=1,...,n\}$, the treatment generative model is updated using
\begin{equation}
\mathcal{L}_X(\phi)
=
-\frac{1}{|\mathcal{B}|}
\sum_{i\in\mathcal{B}}
\log p_\phi(x^{(i)} \mid w^{(i)},\mathbf{z}_0^{(i)}, \mathbf{z}_2^{(i)}).
\label{eq:loss_x_batch}
\end{equation}

(c) Given the current latent states and IV, the outcome generative model is updated using the IV-integrated pseudo-likelihood term, 
\begin{equation}
\mathcal{L}_Y(\omega)
=
-\frac{1}{|\mathcal{B}|}
\sum_{i\in\mathcal{B}}
\log p_{\mathrm{IV}}(y^{(i)} \mid w^{(i)},\mathbf{z}_0^{(i)},\mathbf{z}_1^{(i)},\mathbf{z}_2^{(i)}).
\label{eq:loss_y_batch}
\end{equation}

The pseudo-likelihood term is calculated as:
\begin{equation}
\log p_{\mathrm{IV}}(y^{(i)}\mid w^{(i)},\mathbf{z}^{(i)}_0,\mathbf{z}^{(i)}_1,\mathbf{z}^{(i)}_2)
\approx
\log\left\{
\frac{1}{M}\sum_{m=1}^{M}
p_\omega(y^{(i)}\mid x^{(i,m)},\mathbf{z}^{(i)}_0,\mathbf{z}^{(i)}_1)
\right\},
\label{eq:mc_iv_likelihood}
\end{equation}
where $\{x^{(i,m)}|m=1,...,M\}$ are Monte Carlo samples, $M$ is the Monte Carlo sample size, and $x^{(i,m)}$ is sampled from treatment generative model $x^{(i,m)}\sim p_\phi(x\mid w^{(i)},\mathbf{z}^{(i)}_0,\mathbf{z}^{(i)}_2)$.

(d) Given the current parameters $(\theta,\phi,\omega)$ of generative models, subject-specific latent variables are updated by maximizing the corresponding IV quasi-posterior. For each $i\in\mathcal{B}$, we update $\mathbf{z}^{(i)}$ by gradient ascent on the logarithm of Eq.~\eqref{eq:iv_quasi_posterior},
\begin{equation}
\log p(\mathbf{z}^{(i)})
+
\log p_\theta(\mathbf{v}^{(i)} \mid \mathbf{z}^{(i)})
+
\log p_\phi(x^{(i)} \mid w^{(i)},\mathbf{z}_0^{(i)}, \mathbf{z}_2^{(i)})
+
\gamma\,\log p_{\mathrm{IV}}(y^{(i)} \mid w^{(i)},\mathbf{z}_0^{(i)},\mathbf{z}_1^{(i)},\mathbf{z}_2^{(i)}).
\label{eq:latent_update_objective}
\end{equation}

We draw $K=10$ candidates from independent initializations and retain the one with the smallest IV-moment residual on the training sample (see details in Appendix~\ref{app:baseline_selection}). The complete training procedure of BGM-IV is given in Algorithm~\ref{alg:causalbgm_iv_appendix} in Appendix~\ref{app:detailed_algorithm}.

\subsection{Structural Prediction and Uncertainty Quantification}
\label{sec:model_inference}

After model fitting,  we can estimate the target structural function $g_0(x,\mathbf{v})$. Given any intervention value $x$ and the covariates $\mathbf{v}$ in the inference stage, BGM-IV first infers the latent variable under the \textit{covariate-only} posterior using maximum a posteriori (MAP) estimation,
\begin{equation}
\hat{\mathbf{z}}(\mathbf{v})
=
\arg\max_{\mathbf{z}}
\left\{\log p(\mathbf{z})+\log p_\theta(\mathbf{v}\mid \mathbf{z})\right\}.
\label{eq:test_time_map}
\end{equation}
The structural function is then estimated as
\begin{equation}
\hat g(x,\mathbf{v})=\mu_\omega(x,\hat{\mathbf{z}}_0(\mathbf{v}),\hat{\mathbf{z}}_1(\mathbf{v})),
\label{eq:structural_prediction}
\end{equation}
where $\mu_\omega$ is the mean function of the outcome generator. This step does not condition on $W$, $Y$, or the observed endogenous treatment assignment. $x$ is the intervention value provided to the fitted structural function.

The MAP readout is a point estimate. A key novelty of the generative model is that the same fitted model also quantifies the uncertainty of its predictions. To this end, BGM-IV samples the covariate-only posterior $p(\mathbf z\mid\mathbf v)\propto p(\mathbf z)\,p_\theta(\mathbf v\mid\mathbf z)$ by Hamiltonian Monte Carlo with several overdispersed chains. Each draw $\mathbf z^{(j)}$ yields a structural value $\mu_\omega(x,\mathbf z_0^{(j)},\mathbf z_1^{(j)})$ and an outcome distribution $\mathcal N\big(\mu_\omega(x,\mathbf z_0^{(j)},\mathbf z_1^{(j)}),\,\sigma_\omega^2(x,\mathbf z_0^{(j)},\mathbf z_1^{(j)})\big)$. Averaging the latter over the $J$ draws gives the model's posterior predictive distribution at $(x,\mathbf v)$,
\begin{equation}
\hat p(y\mid x,\mathbf v)
=
\frac{1}{J}\sum_{j=1}^{J}
\mathcal N\!\big(y;\,\mu_\omega(x,\mathbf z_0^{(j)},\mathbf z_1^{(j)}),\,\sigma_\omega^2(x,\mathbf z_0^{(j)},\mathbf z_1^{(j)})\big),
\label{eq:posterior_predictive}
\end{equation}
whose quantiles define predictive intervals. These predictive intervals propagate uncertainty in the latent state and incorporate the component variance $\sigma_\omega^2$ of the fitted model. Throughout the paper, we report MAP point estimates and use the predictive distribution for interval estimates.

\section{Experiments}
\label{sec:experiments}

\subsection{Experimental Setup}
\label{sec:experimental_setup}

\textbf{Benchmarks.} We evaluate BGM-IV on three benchmarks of increasing covariate complexity. The first is the demand-design benchmark derived from the airline-pricing simulation, which has low-dimensional covariates and has been widely used by existing IV methods~\citep{hartford2017deepiv,shao2024dmliv,xu2021dfiv}. The other two are high-dimensional benchmarks, in which the covariate variable is observed through a 1-D high-dimensional vector or through a 2-D image. The structural function is known in all three benchmarks, so that both the point estimates and the predictive intervals can be evaluated against the truth. The full details of the data-generating processes are given in Appendix~\ref{app:experiment_details}.

\textbf{Baselines.} We compare BGM-IV with five nonlinear IV methods, including KIV~\citep{singh2019kiv}, DeepIV~\citep{hartford2017deepiv}, DFIV~\citep{xu2021dfiv}, DML-IV~\citep{shao2024dmliv}, and FIL~\citep{wang2022fil}, using their publicly released implementations and default hyperparameters. Every method except KIV, which has no random initialization, is trained from ten independent initializations per repetition, and one run is retained by that method's own training-side criterion to maximize the baseline performance. The baseline implementations and their selection criteria are described in Appendix~\ref{app:baseline_selection}, and the implementation details of BGM-IV are given in Appendix~\ref{app:bgm_iv_hyperparameters}.

\textbf{Evaluation metrics.} For point estimation, we report the structural mean squared error (MSE) on the original outcome scale, computed on the evaluation grid of prior demand-design benchmarks. Each configuration is repeated 20 times with independent training samples shared by all methods to ensure fair comparison, and paired comparisons with the strongest baseline on the two high-dimensional benchmarks are reported in Appendix~\ref{app:pvalue_comparisons}. For uncertainty quantification, we report the coverage of new outcomes and the width of the predictive intervals of BGM-IV on the same grid, whose evaluation is detailed in Appendix~\ref{app:baseline_selection}. All baseline implementations return point estimates only.

\subsection{Demand Design Experiments}
\label{sec:demand_design}

The demand-design benchmark models counterfactual demand estimation in airline pricing with low-dimensional covariates. The treatment $P$ is the ticket price, the outcome $Y$ is demand, and the observed covariates are time of year $T$ and customer group $S$, which induce heterogeneous price sensitivity. Prices are endogenous because an unobserved demand shock affects both the price and the realized demand, with the strength of this confounding set by a correlation parameter $\rho$. The instrument is a cost shifter that changes the price but is excluded from the demand equation. The structural function is $f_0(P,T,S)=100+(10+P)S\psi(T)-2P$ with a nonlinear seasonal function $\psi$, and the full design is given in Appendix~\ref{app:experiment_details}. We vary the training sample size $n\in\{1{,}000,5{,}000,10{,}000\}$ and the confounding level $\rho\in\{0.1,0.25,0.5,0.75,0.9\}$.

Figure~\ref{fig:demand_lowdim_box} summarizes the results and the complete values are in Table~\ref{tab:demand_mse} of Appendix~\ref{app:full_results}. BGM-IV attains the lowest mean structural MSE in all 15 configurations. For example, when the sample size is $n=1{,}000$, BGM-IV reduces the MSE by $57.9\%$-$62.8\%$ relative to the strongest baseline, while exhibiting $2.7$-fold to $7.1$-fold lower variability across repetitions. Taken together, these results show the superiority of BGM-IV in accurate and robust structural estimates on this widely used classical IV benchmark.  

 Section~\ref{sec:quasi_posterior} argues that the latent states used for prediction are inferred from the covariates alone and therefore preserve conditional mean exogeneity. The unobserved confounder and the structural error are explicit in the demand design experiment, which allows this argument to be checked empirically. We run the check at the strongest confounding level $\rho=0.9$,  where the unobserved confounder accounts for $81\%$ of the variance of the structural error and a latent state that absorbed this error would reintroduce the most endogeneity. The latent states show no detectable association with the structural error or with the unobserved confounder, on training and on held-out data. Appendix~\ref{app:latent_exogeneity} provides details, including the negative and positive controls, and the complete results.

\begin{figure}[t]
  \centering
  \includegraphics[width=0.97\linewidth]{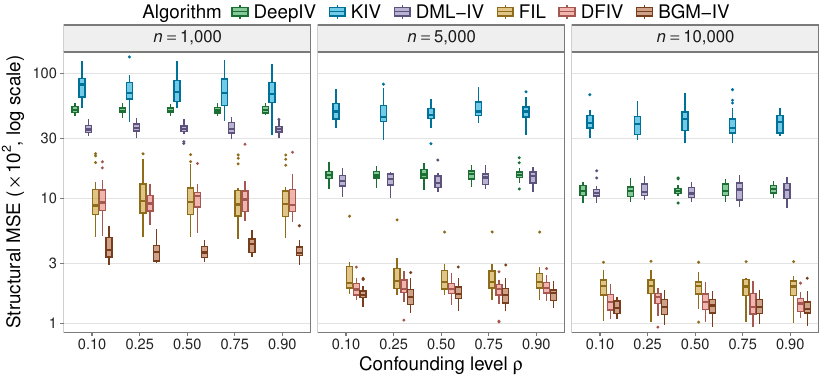}
  \caption{Structural MSE on the demand-design benchmark by sample size and confounding level.}
  \label{fig:demand_lowdim_box}
\vspace{-15pt}
\end{figure}

\subsection{High-Dimensional Vector-Proxy Experiments}
\label{sec:vector_proxy}

The vector-proxy benchmark is the high-dimensional extension of the demand design experiment. The structural equation, price equation, and instrument remain unchanged, but the customer group is no longer observed as a discrete label. Instead, each group $s\in\{1,\ldots,7\}$ has a prototype $\boldsymbol{\mu}_s\in\mathbb{R}^{784}$ drawn from $\mathcal N(\mathbf 0,\mathbf I)$, and we observe the noisy proxy $\mathbf{r}=\boldsymbol{\mu}_S+\sigma_{\mathrm{rep}}\boldsymbol{\eta}$ with $\boldsymbol{\eta}\sim\mathcal N(\mathbf 0,\mathbf I)$. This construction follows the proxy-variable view that latent causal factors may be observed only through noisy measurements~\citep{kuroki2014measurement,miao2018proxy}, but here the proxy mechanism is fully controlled so the benchmark isolates representation learning rather than identification by proxy variables. To control how much of the proxy each method has to digest, every method receives the covariate $(T,\mathrm{PCA}_k(\mathbf r))$, the first $k\in\{1,2,4,6,8,16,32,64,128\}$ principal components fitted on the training proxies, and keeps its demand-design architecture and hyperparameters. We fix $\rho=0.5$ and vary $n\in\{5{,}000,10{,}000,20{,}000,50{,}000\}$.

\begin{figure}[t]
  \centering
  \includegraphics[width=\linewidth]{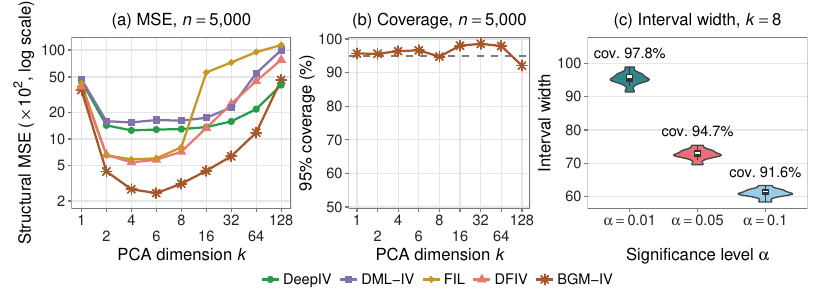}
  \caption{Vector-proxy benchmark at $n=5{,}000$. (a) Structural MSE against the number $k$ of principal components. (b) Coverage of the 95\% predictive intervals of BGM-IV, the dashed line is the nominal level. (c) Widths of the 99\%, 95\%, and 90\% predictive intervals at $k=8$ over repetitions with the mean coverage rate.}
  \label{fig:vector_mse_calibration}
  \vspace{-10pt}
\end{figure}

Figure~\ref{fig:vector_mse_calibration} shows the results at $n=5{,}000$. The complete values for all sample sizes are in Table~\ref{tab:vector_mse} of Appendix~\ref{app:full_results}. Across all sample sizes, BGM-IV achieves the lowest mean MSE in 35 of the 36 configurations. With a single component with limited information to separate the seven groups, all methods perform poorly and BGM-IV still has the lowest mean MSE, and for $k\ge2$ its improvement over the strongest baseline is statistically significant in 30 of the 32 configurations (Table~\ref{tab:paired_vector} of Appendix~\ref{app:pvalue_comparisons}). For instance, when $k=2$-$8$, it reduces structural MSE by 32\%-58\% relative to the strongest baseline. For $k>8$, where the additional components primarily introduce nuisance variation in this design, BGM-IV retains the lowest mean error for every $k$ when $n\ge10{,}000$, while several competing methods deteriorate sharply as additional nuisance dimensions are included. 

Beyond point estimation, BGM-IV uniquely quantifies the uncertainty of its predictions. Panel (b) shows that the coverage of the 95\% predictive intervals remains near the nominal level for every $k$, and panel (c) shows that at $k=8$ the 90\%, 95\%, and 99\% intervals all attain close to nominal coverage, with widths that increase with the nominal coverage level $1-\alpha$.

\subsection{High-Dimensional Image Experiments}
\label{sec:image_experiments}

The image-covariate benchmark follows the high-dimensional demand setting introduced by DeepIV~\citep{hartford2017deepiv}, which has since been adopted as a standard test of nonlinear IV methods with image covariates~\citep{xu2021dfiv,shao2024dmliv}. The causal mechanism of the demand design is unchanged, but the customer group is no longer given as a label. It is observed only through a $28\times28$ handwritten-digit image from MNIST~\citep{lecun1998mnist} whose digit equals the group, so the observed covariates are $T$ and a 784-dimensional image, and a method has to recover the group from raw pixels in order to estimate the structural function. We use the sample sizes and confounding levels of Section~\ref{sec:demand_design}.

\begin{wrapfigure}{r}{0.44\linewidth}
  \vspace{-25pt}
  \centering
  \includegraphics[width=\linewidth]{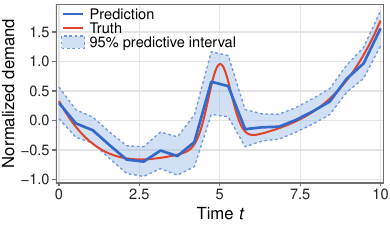}
  \caption{A slice of the demand curve estimated by BGM-IV on the image-covariate benchmark ($n=50{,}000$, $\rho=0.5$).}
  \label{fig:mnist_curve}
  \vspace{-15pt}
\end{wrapfigure}

Table~\ref{tab:mnist_main} reports the complete benchmark results. BGM-IV attains the lowest mean MSE in all 15 configurations, and each improvement over the strongest baseline is statistically significant (Table~\ref{tab:paired_mnist} of Appendix~\ref{app:pvalue_comparisons}). Its advantage grows with the sample size: relative to the strongest baseline, its MSE is about 1.3 times smaller at $n=1{,}000$, 2.6 to 2.9 times smaller at $n=5{,}000$, and more than three times smaller at $n=10{,}000$. These
results show the superiority of BGM-IV in accurate and robust structural estimates on this widely used high-dimensional IV benchmark.

Figure~\ref{fig:mnist_curve} illustrates the unique uncertainty quantification of BGM-IV on this benchmark with a model trained on $n=50{,}000$ samples at $\rho=0.5$. It shows a slice of the estimated demand curve with its 95\% predictive band. Over 20 repetitions of this setting, the 90\%, 95\%, and 99\% predictive intervals cover 90.7\%, 93.8\%, and 96.8\% of new outcomes.

In addition, building on the image-covariate benchmark, we consider the case in which the instrument is only weakly correlated with the treatment in Appendix~\ref{app:full_results_weakiv}, where BGM-IV again attains the lowest MSE among all methods at every instrument strength and sample size considered.

\begin{table}[t]
\vspace{-10pt}
\centering
\caption{Structural MSE on the image-covariate benchmark (entries divided by $10^{2}$), mean $\pm$ standard deviation over 20 repetitions. The lowest mean in each row is in bold.}
\label{tab:mnist_main}
\footnotesize
\setlength{\tabcolsep}{2pt}
\renewcommand{\arraystretch}{0.9}
\begin{tabular}{@{}cc cccccc@{}}
\toprule
$n$ & $\rho$ & DeepIV & KIV & DML-IV & FIL & DFIV & \textbf{BGM-IV} \\
\midrule
\multirow{5}{*}{$1{,}000$} & 0.1 & 153.21 $\pm$ 6.13 & 208.31 $\pm$ 2.45 & 133.32 $\pm$ 2.66 & 131.51 $\pm$ 13.87 & 102.65 $\pm$ 5.72 & \textbf{80.04 $\pm$ 15.05} \\
 & 0.25 & 155.08 $\pm$ 5.31 & 207.54 $\pm$ 2.76 & 132.71 $\pm$ 3.91 & 133.82 $\pm$ 17.40 & 102.92 $\pm$ 5.71 & \textbf{79.18 $\pm$ 11.15} \\
 & 0.5 & 155.46 $\pm$ 5.84 & 208.57 $\pm$ 2.62 & 133.54 $\pm$ 3.98 & 133.48 $\pm$ 16.94 & 103.20 $\pm$ 4.47 & \textbf{75.85 $\pm$ 11.54} \\
 & 0.75 & 153.93 $\pm$ 5.25 & 209.06 $\pm$ 2.88 & 133.21 $\pm$ 3.93 & 133.16 $\pm$ 16.65 & 102.90 $\pm$ 5.85 & \textbf{80.50 $\pm$ 8.48} \\
 & 0.9 & 154.11 $\pm$ 5.40 & 207.33 $\pm$ 2.32 & 132.23 $\pm$ 4.35 & 133.70 $\pm$ 16.89 & 101.58 $\pm$ 4.71 & \textbf{78.36 $\pm$ 10.16} \\
\midrule
\multirow{5}{*}{$5{,}000$} & 0.1 & 85.81 $\pm$ 9.16 & 192.53 $\pm$ 2.21 & 75.51 $\pm$ 3.74 & 89.70 $\pm$ 12.85 & 69.11 $\pm$ 4.07 & \textbf{26.50 $\pm$ 4.59} \\
 & 0.25 & 84.40 $\pm$ 7.25 & 192.19 $\pm$ 2.21 & 76.93 $\pm$ 3.00 & 89.75 $\pm$ 12.91 & 71.22 $\pm$ 1.93 & \textbf{25.32 $\pm$ 2.62} \\
 & 0.5 & 85.04 $\pm$ 8.80 & 192.47 $\pm$ 2.19 & 75.13 $\pm$ 2.30 & 91.08 $\pm$ 13.52 & 71.10 $\pm$ 3.46 & \textbf{25.54 $\pm$ 2.01} \\
 & 0.75 & 84.66 $\pm$ 7.48 & 192.05 $\pm$ 1.91 & 75.58 $\pm$ 2.85 & 91.20 $\pm$ 13.04 & 71.12 $\pm$ 4.00 & \textbf{24.91 $\pm$ 2.13} \\
 & 0.9 & 87.67 $\pm$ 7.60 & 192.22 $\pm$ 1.78 & 76.15 $\pm$ 3.11 & 91.19 $\pm$ 12.98 & 69.33 $\pm$ 4.58 & \textbf{25.94 $\pm$ 3.57} \\
\midrule
\multirow{5}{*}{$10{,}000$} & 0.1 & 80.57 $\pm$ 6.44 & 187.57 $\pm$ 1.80 & 63.76 $\pm$ 2.06 & 78.08 $\pm$ 19.88 & 55.38 $\pm$ 5.92 & \textbf{17.12 $\pm$ 1.64} \\
 & 0.25 & 79.75 $\pm$ 7.40 & 187.63 $\pm$ 1.39 & 63.87 $\pm$ 2.40 & 73.48 $\pm$ 7.70 & 55.82 $\pm$ 6.06 & \textbf{17.16 $\pm$ 1.24} \\
 & 0.5 & 78.48 $\pm$ 6.22 & 187.20 $\pm$ 1.73 & 63.75 $\pm$ 2.89 & 79.24 $\pm$ 19.30 & 55.73 $\pm$ 6.37 & \textbf{17.14 $\pm$ 1.39} \\
 & 0.75 & 77.96 $\pm$ 5.53 & 187.31 $\pm$ 1.72 & 63.69 $\pm$ 2.27 & 79.91 $\pm$ 19.75 & 55.68 $\pm$ 5.70 & \textbf{17.13 $\pm$ 1.25} \\
 & 0.9 & 80.88 $\pm$ 6.45 & 187.63 $\pm$ 2.00 & 63.68 $\pm$ 2.03 & 79.77 $\pm$ 19.98 & 54.69 $\pm$ 6.51 & \textbf{17.14 $\pm$ 1.84} \\
\bottomrule
\end{tabular}
\vspace{-15pt}
\end{table}

\subsection{Ablation Studies}
\label{sec:ablation}

We examine the three components of BGM-IV on the demand design at $n=1{,}000$ with three settings: the standard design, a design in which the structural disturbance is amplified by a factor of 100, and a design with five additional nuisance covariates. Each variant changes one component and keeps every other setting unchanged, so that it is paired with the full model by repetition. Appendix~\ref{app:ablation} describes the settings and variants and compares BGM-IV with the strongest baselines in all three settings, and Table~\ref{tab:ablation_main} reports the results.

\textbf{IV-aware EGM initialization.} Without the IV-aware EGM initialization, the structural MSE is 17.7 to 18.9 times larger than that of the full model, and it is larger in every one of the 60 paired repetitions. These results indicate that the initialization plays a critical role in optimization.

\textbf{IV-integrated pseudo-likelihood.} Under amplified confounding, replacing the IV-integrated pseudo-likelihood by the likelihood at the observed treatment increases the structural MSE by a factor of 1.18, 1.98, and 3.63 at $\rho=0.1$, $0.5$, and $0.9$. The increase grows with the confounding level because the observed-treatment likelihood fits the outcome at the endogenous treatment, and the resulting bias grows with the correlation $\rho$ between the confounder and the structural disturbance.

\textbf{Structured latent space.} With five nuisance covariates and a partition specified from the design, replacing the latent partition by an unstructured latent space increases the structural MSE by a factor of 1.40 to 1.44. A regression of the nuisance covariates on the trained latent states shows that they are captured by the nuisance block and not by the treatment and outcome generators (Appendix~\ref{app:ablation}).

\begin{table}[t]
\centering
\caption{Ablation study on the demand design at $n=1{,}000$ (structural MSE divided by $10^{2}$, mean $\pm$ standard deviation over 20 repetitions). A check mark indicates that the component is kept: the EGM initialization, the IV-integrated pseudo-likelihood (IV-PL), and the latent partition.}
\label{tab:ablation_main}
\footnotesize
\setlength{\tabcolsep}{5pt}
\begin{tabular}{@{}l ccc ccc@{}}
\toprule
Setting & EGM init. & IV-PL & Partition & $\rho=0.1$ & $\rho=0.5$ & $\rho=0.9$ \\
\midrule
\multirow{2}{*}{Standard} & \checkmark & \checkmark & \checkmark & \textbf{4.15 $\pm$ 0.91} & \textbf{3.74 $\pm$ 0.47} & \textbf{3.87 $\pm$ 0.69} \\
 & $\times$ & \checkmark & \checkmark & 73.39 $\pm$ 8.63 & 70.56 $\pm$ 7.13 & 69.14 $\pm$ 7.29 \\
\midrule
\multirow{2}{*}{Amplified confounding} & \checkmark & \checkmark & \checkmark & \textbf{35.21 $\pm$ 8.03} & \textbf{33.82 $\pm$ 9.01} & \textbf{31.65 $\pm$ 6.23} \\
 & \checkmark & $\times$ & \checkmark & 41.63 $\pm$ 9.75 & 67.09 $\pm$ 19.91 & 114.77 $\pm$ 36.21 \\
\midrule
\multirow{2}{*}{Nuisance covariates} & \checkmark & \checkmark & \checkmark & \textbf{32.34 $\pm$ 9.98} & \textbf{32.86 $\pm$ 9.89} & \textbf{30.53 $\pm$ 8.69} \\
 & \checkmark & \checkmark & $\times$ & 45.16 $\pm$ 4.50 & 47.16 $\pm$ 4.93 & 43.73 $\pm$ 4.84 \\
\bottomrule
\end{tabular}
\vspace{-15pt}
\end{table}

\section{Conclusion}

We introduced BGM-IV, an AI-powered Bayesian generative framework for nonlinear IV regression with high-dimensional covariates. By combining a causally structured latent representation with an IV-integrated pseudo-likelihood, BGM-IV learns structural effects from instrument-induced treatment variation and provides both point estimates and uncertainty quantification. Across diverse benchmarks, it achieves lower structural MSE in nearly all evaluated settings, with the largest gains under high-dimensional covariates, while attaining near-nominal predictive coverage. Computational cost is one of the limitations (see Appendix~\ref{app:computational_cost}). Future work will extend BGM-IV to more complex treatment or instrument settings, such as longitudinal or dynamic regimes.

\subsubsection*{Acknowledgments}
G.Y.L. and Q.L. were supported in part by the National Human Genome Research Institute (NHGRI) under Award Number R00HG013661, Operations Core of the Claude D. Pepper Older Americans Independence Center at Yale School of Medicine (P30AG021342), and the YSPH Transformation Pilot Grant. H.L. is supported by the National Institute on Drug Abuse(NIDA) under Award Number R00DA057487. 

\bibliography{iclr2027_conference}

@article{liu2026missingness,
  title={Missingness-aware Data Imputation via AI-powered Bayesian Generative Modeling},
  author={Liu, Qiao},
  journal={arXiv preprint arXiv:2605.01676},
  year={2026}
}

@article{liu2024encoding,
  title={An encoding generative modeling approach to dimension reduction and covariate adjustment in causal inference with observational studies},
  author={Liu, Qiao and Chen, Zhongren and Wong, Wing Hung},
  journal={Proceedings of the National Academy of Sciences},
  volume={121},
  number={23},
  pages={e2322376121},
  year={2024},
  publisher={National Academy of Sciences}
}

@article{liu2026causalbgm,
author = {Qiao Liu and Wing Hung Wong},
title = {An AI-powered Bayesian Generative Modeling Approach for Causal Inference in Observational Studies},
journal = {Journal of the American Statistical Association},
volume = {121},
number = {554},
pages = {976--987},
year = {2026},
publisher = {Taylor \& Francis},
doi = {10.1080/01621459.2026.2654227},


URL = { 
    
        https://doi.org/10.1080/01621459.2026.2654227
    
    

},
eprint = { 
    
        https://doi.org/10.1080/01621459.2026.2654227
    
    

}

}

@misc{liu2026bgm,
      title={An AI-powered Bayesian Generative Modeling Approach for Arbitrary Conditional Inference}, 
      author={Qiao Liu and Wing Hung Wong},
      year={2026},
      eprint={2601.05355},
      archivePrefix={arXiv},
      primaryClass={stat.ML},
      url={https://arxiv.org/abs/2601.05355}, 
}

@article{angrist1996iv,
  title={Identification of causal effects using instrumental variables},
  author={Angrist, Joshua D and Imbens, Guido W and Rubin, Donald B},
  journal={Journal of the American statistical Association},
  volume={91},
  number={434},
  pages={444--455},
  year={1996},
  publisher={Taylor \& Francis}
}

@article{stock2003ivhistory,
  title={Retrospectives: Who invented instrumental variable regression?},
  author={Stock, James H and Trebbi, Francesco},
  journal={Journal of Economic Perspectives},
  volume={17},
  number={3},
  pages={177--194},
  year={2003},
  publisher={American Economic Association}
}

@book{pearl2009causality, place={Cambridge}, edition={2}, title={Causality}, publisher={Cambridge University Press}, author={Pearl, Judea}, year={2009}}

@book{imbens2015causal, place={Cambridge}, title={Causal Inference for Statistics, Social, and Biomedical Sciences: An Introduction}, publisher={Cambridge University Press}, author={Imbens, Guido W. and Rubin, Donald B.}, year={2015}}

@book{hernan2025causal,
  title={Causal Inference: What If},
  author={Hernan, M.A. and Robins, J.M.},
  isbn={9781420076165},
  lccn={2022050839},
  series={Chapman \& Hall/CRC Monographs on Statistics \& Applied Probab},
  url={https://books.google.com/books?id=_KnHIAAACAAJ},
  year={2020},
  publisher={CRC Press}
}

@book{angrist2009mostly,
  title={Mostly harmless econometrics: An empiricist's companion},
  author={Angrist, Joshua D and Pischke, J{\"o}rn-Steffen},
  year={2009},
  publisher={Princeton university press}
}

@article{newey2003nonparametric,
  title={Instrumental variable estimation of nonparametric models},
  author={Newey, Whitney K and Powell, James L},
  journal={Econometrica},
  volume={71},
  number={5},
  pages={1565--1578},
  year={2003},
  publisher={Wiley Online Library}
}

@article{blundell2007semi,
  title={Semi-nonparametric IV estimation of shape-invariant Engel curves},
  author={Blundell, Richard and Chen, Xiaohong and Kristensen, Dennis},
  journal={Econometrica},
  volume={75},
  number={6},
  pages={1613--1669},
  year={2007},
  publisher={Wiley Online Library}
}

@article{darolles2011nonparametric,
  title={Nonparametric instrumental regression},
  author={Darolles, Serge and Fan, Yanqin and Florens, Jean-Pierre and Renault, Eric},
  journal={Econometrica},
  volume={79},
  number={5},
  pages={1541--1565},
  year={2011},
  publisher={Wiley Online Library}
}

@article{chen2012conditional,
  title={Estimation of nonparametric conditional moment models with possibly nonsmooth generalized residuals},
  author={Chen, Xiaohong and Pouzo, Demian},
  journal={Econometrica},
  volume={80},
  number={1},
  pages={277--321},
  year={2012},
  publisher={Wiley Online Library}
}

@article{hansen1982gmm,
 ISSN = {00129682, 14680262},
 URL = {http://www.jstor.org/stable/1912775},
 author = {Lars Peter Hansen},
 journal = {Econometrica},
 number = {4},
 pages = {1029--1054},
 publisher = {[Wiley, Econometric Society]},
 title = {Large Sample Properties of Generalized Method of Moments Estimators},
 urldate = {2026-09-26},
 volume = {50},
 year = {1982}
}

@article{singh2019kiv,
  title={Kernel instrumental variable regression},
  author={Singh, Rahul and Sahani, Maneesh and Gretton, Arthur},
  journal={Advances in Neural Information Processing Systems},
  volume={32},
  year={2019}
}

@article{damour2021overlap,
  title={Overlap in observational studies with high-dimensional covariates},
  author={D’Amour, Alexander and Ding, Peng and Feller, Avi and Lei, Lihua and Sekhon, Jasjeet},
  journal={Journal of Econometrics},
  volume={221},
  number={2},
  pages={644--654},
  year={2021},
  publisher={Elsevier}
}

@article{chernozhukov2018dml,
    author = {Chernozhukov, Victor and Chetverikov, Denis and Demirer, Mert and Duflo, Esther and Hansen, Christian and Newey, Whitney and Robins, James},
    title = {Double/debiased machine learning for treatment and structural parameters},
    journal = {The Econometrics Journal},
    volume = {21},
    number = {1},
    pages = {C1-C68},
    year = {2018},
    month = {02},
    issn = {1368-4221},
    doi = {10.1111/ectj.12097},
    url = {https://doi.org/10.1111/ectj.12097},
    eprint = {https://academic.oup.com/ectj/article-pdf/21/1/C1/27684918/ectj00c1.pdf},
}

@inproceedings{hartford2017deepiv,
  title={Deep IV: A flexible approach for counterfactual prediction},
  author={Hartford, Jason and Lewis, Greg and Leyton-Brown, Kevin and Taddy, Matt},
  booktitle={International conference on machine learning},
  pages={1414--1423},
  year={2017},
  organization={PMLR}
}

@inproceedings{
xu2021dfiv,
title={Learning Deep Features in Instrumental Variable Regression},
author={Liyuan Xu and Yutian Chen and Siddarth Srinivasan and Nando de Freitas and Arnaud Doucet and Arthur Gretton},
booktitle={International Conference on Learning Representations},
year={2021},
url={https://openreview.net/forum?id=sy4Kg_ZQmS7}
}

@article{bennett2019deepgmm,
  title={Deep generalized method of moments for instrumental variable analysis},
  author={Bennett, Andrew and Kallus, Nathan and Schnabel, Tobias},
  journal={Advances in neural information processing systems},
  volume={32},
  year={2019}
}

@ARTICLE{lecun1998mnist,
  author={LeCun, Y. and Bottou, L. and Bengio, Y. and Haffner, P.},
  journal={Proceedings of the IEEE}, 
  title={Gradient-based learning applied to document recognition}, 
  year={1998},
  volume={86},
  number={11},
  pages={2278-2324},
  doi={10.1109/5.726791}}

@article{paszke2019pytorch,
  title={Pytorch: An imperative style, high-performance deep learning library},
  author={Paszke, Adam and Gross, Sam and Massa, Francisco and Lerer, Adam and Bradbury, James and Chanan, Gregory and Killeen, Trevor and Lin, Zeming and Gimelshein, Natalia and Antiga, Luca and others},
  journal={Advances in neural information processing systems},
  volume={32},
  year={2019}
}

@article{kuroki2014measurement,
  title={Measurement bias and effect restoration in causal inference},
  author={Kuroki, Manabu and Pearl, Judea},
  journal={Biometrika},
  pages={423--437},
  year={2014},
  publisher={JSTOR}
}

@article{miao2018proxy,
  title={Identifying causal effects with proxy variables of an unmeasured confounder},
  author={Miao, Wang and Geng, Zhi and Tchetgen Tchetgen, Eric J},
  journal={Biometrika},
  volume={105},
  number={4},
  pages={987--993},
  year={2018},
  publisher={Oxford University Press}
}

@inproceedings{johansson2016learning,
  title={Learning representations for counterfactual inference},
  author={Johansson, Fredrik and Shalit, Uri and Sontag, David},
  booktitle={International conference on machine learning},
  pages={3020--3029},
  year={2016},
  organization={PMLR}
}

@article{louizos2017causal,
  title={Causal effect inference with deep latent-variable models},
  author={Louizos, Christos and Shalit, Uri and Mooij, Joris M and Sontag, David and Zemel, Richard and Welling, Max},
  journal={Advances in neural information processing systems},
  volume={30},
  year={2017}
}

@article{hall2005nonparametric,
author = {Peter Hall and Joel L. Horowitz},
title = {{Nonparametric methods for inference in the presence of instrumental variables}},
volume = {33},
journal = {The Annals of Statistics},
number = {6},
publisher = {Institute of Mathematical Statistics},
pages = {2904 -- 2929},
year = {2005},
doi = {10.1214/009053605000000714},
URL = {https://doi.org/10.1214/009053605000000714}
}

@article{carrasco2007linear,
  title={Linear inverse problems in structural econometrics estimation based on spectral decomposition and regularization},
  author={Carrasco, Marine and Florens, Jean-Pierre and Renault, Eric},
  journal={Handbook of econometrics},
  volume={6},
  pages={5633--5751},
  year={2007},
  publisher={Elsevier}
}

@article{angrist1995identification,
 ISSN = {00129682, 14680262},
 URL = {http://www.jstor.org/stable/2951620},
 author = {Guido W. Imbens and Joshua D. Angrist},
 journal = {Econometrica},
 number = {2},
 pages = {467--475},
 publisher = {[Wiley, Econometric Society]},
 title = {Identification and Estimation of Local Average Treatment Effects},
 urldate = {2026-09-17},
 volume = {62},
 year = {1994}
}

@article{bound1995problems,
  title={Problems with instrumental variables estimation when the correlation between the instruments and the endogenous explanatory variable is weak},
  author={Bound, John and Jaeger, David A and Baker, Regina M},
  journal={Journal of the American statistical association},
  volume={90},
  number={430},
  pages={443--450},
  year={1995},
  publisher={Taylor \& Francis}
}

@article{chernozhukov2022automatic,
  title={Automatic debiased machine learning of causal and structural effects},
  author={Chernozhukov, Victor and Newey, Whitney K and Singh, Rahul},
  journal={Econometrica},
  volume={90},
  number={3},
  pages={967--1027},
  year={2022},
  publisher={Wiley Online Library}
}

@inproceedings{shalit2017estimating,
  title={Estimating individual treatment effect: generalization bounds and algorithms},
  author={Shalit, Uri and Johansson, Fredrik D and Sontag, David},
  booktitle={International conference on machine learning},
  pages={3076--3085},
  year={2017},
  organization={PMLR}
}

@article{scholkopf2021toward,
  title={Toward causal representation learning},
  author={Sch{\"o}lkopf, Bernhard and Locatello, Francesco and Bauer, Stefan and Ke, Nan Rosemary and Kalchbrenner, Nal and Goyal, Anirudh and Bengio, Yoshua},
  journal={Proceedings of the IEEE},
  volume={109},
  number={5},
  pages={612--634},
  year={2021},
  publisher={IEEE}
}

@article{dikkala2020minimax,
  title={Minimax estimation of conditional moment models},
  author={Dikkala, Nishanth and Lewis, Greg and Mackey, Lester and Syrgkanis, Vasilis},
  journal={Advances in Neural Information Processing Systems},
  volume={33},
  pages={12248--12262},
  year={2020}
}

@article{d2011completeness,
  title={On the completeness condition in nonparametric instrumental problems},
  author={D’Haultfoeuille, Xavier},
  journal={Econometric Theory},
  volume={27},
  number={3},
  pages={460--471},
  year={2011},
  publisher={Cambridge University Press}
}

@article{wu2025instrumental,
  title={Instrumental variables in causal inference and machine learning: A survey},
  author={Wu, Anpeng and Kuang, Kun and Xiong, Ruoxuan and Wu, Fei},
  journal={ACM Computing Surveys},
  volume={57},
  number={11},
  pages={1--36},
  year={2025},
  publisher={ACM New York, NY}
}

@article{Hernán02012019,
author = {Miguel A. Hernán and John Hsu and Brian Healy},
title = {A Second Chance to Get Causal Inference Right: A Classification of Data Science Tasks},
journal = {CHANCE},
volume = {32},
number = {1},
pages = {42--49},
year = {2019},
publisher = {Taylor \& Francis},
doi = {10.1080/09332480.2019.1579578},


URL = { 
    
        https://doi.org/10.1080/09332480.2019.1579578
    
    

},
eprint = { 
    
        https://doi.org/10.1080/09332480.2019.1579578
    
    

}

}

@article{https://doi.org/10.1111/rssb.12158,
author = {Bissiri, P. G. and Holmes, C. C. and Walker, S. G.},
title = {A general framework for updating belief distributions},
journal = {Journal of the Royal Statistical Society: Series B (Statistical Methodology)},
volume = {78},
number = {5},
pages = {1103-1130},
doi = {https://doi.org/10.1111/rssb.12158},
url = {https://rss.onlinelibrary.wiley.com/doi/abs/10.1111/rssb.12158},
eprint = {https://rss.onlinelibrary.wiley.com/doi/pdf/10.1111/rssb.12158},
year = {2016}
}

@book{hyndman2014forecasting,
  title={Forecasting: principles and practice},
  author={Hyndman, Rob J and Athanasopoulos, George and others},
  year={2014},
  publisher={OTexts Melbourne, Australia}
}

@article{holm1979simple,
  title={A simple sequentially rejective multiple test procedure},
  author={Holm, Sture},
  journal={Scandinavian journal of statistics},
  pages={65--70},
  year={1979},
  publisher={JSTOR}
}

@article{ WOS:000277071500016,
Author = {Lipsitch, Marc and Tchetgen, Eric Tchetgen and Cohen, Ted},
Title = {Negative Controls: A Tool for Detecting Confounding and Bias in
   Observational Studies},
Journal = {EPIDEMIOLOGY},
Year = {2010},
Volume = {21},
Number = {3},
Pages = {383-388},
Month = {MAY},
DOI = {10.1097/EDE.0b013e3181d61eeb},
ISSN = {1044-3983},
EISSN = {1531-5487},
ResearcherID-Numbers = {Tchetgen Tchetgen, Eric/AAF-6160-2021
   Lipsitch, Marc/AGD-6299-2022
   Cohen, Ted/G-8101-2011},
ORCID-Numbers = {Lipsitch, Marc/0000-0003-1504-9213
   Cohen, Ted/0000-0002-8091-7198},
Unique-ID = {WOS:000277071500016},
}

@InProceedings{shao2024dmliv,
  title = 	 {Learning Decision Policies with Instrumental Variables through Double Machine Learning},
  author =       {Shao, Daqian and Soleymani, Ashkan and Quinzan, Francesco and Kwiatkowska, Marta},
  booktitle = 	 {Proceedings of the 41st International Conference on Machine Learning},
  pages = 	 {44489--44514},
  year = 	 {2024},
  editor = 	 {Salakhutdinov, Ruslan and Kolter, Zico and Heller, Katherine and Weller, Adrian and Oliver, Nuria and Scarlett, Jonathan and Berkenkamp, Felix},
  volume = 	 {235},
  series = 	 {Proceedings of Machine Learning Research},
  month = 	 {21--27 Jul},
  publisher =    {PMLR},
  url = 	 {https://proceedings.mlr.press/v235/shao24d.html}
}

@inproceedings{wang2022fil,
author = {Wang, Ziyu and Zhou, Yuhao and Zhu, Jun},
title = {Fast instrument learning with faster rates},
year = {2022},
isbn = {9781713871088},
publisher = {Curran Associates Inc.},
address = {Red Hook, NY, USA},
booktitle = {Proceedings of the 36th International Conference on Neural Information Processing Systems},
articleno = {1207},
numpages = {16},
location = {New Orleans, LA, USA},
series = {NIPS '22}
}
\bibliographystyle{iclr2027_conference}


\appendix
\makeatletter
\setlength{\@fptop}{0pt}
\makeatother

\FloatBarrier
\section{Structural-Mean Identification and Conditional Consistency}
\label{app:iv_population_justification}

This appendix studies the structural mean. We establish conditional mean exogeneity for covariate-only readouts and identify the structural function under representation sufficiency and completeness. We characterize the population target of the integrated EGM outcome component, then bound the final selected predictor's IV-moment risk and derive consistency under explicit approximation, error-control and inverse-stability conditions. A complementary observed-data comparison gives another sufficient condition for the selected predictor.

\subsection{Covariate Readouts and Mean Identification}

Write
\[
A=(Z_0,Z_1,Z_2),\qquad A_Y=(Z_0,Z_1),\qquad A_X=(Z_0,Z_2).
\]
For the fixed-representation population arguments, $A=\varphi(V)$ is a measurable map. Representation sufficiency is a separate condition: the structural target must admit $g_0(x,V)=m_0(x,A_Y)$ on the identified support. Measurability alone does not ensure this equality. Let $\lambda_X$ denote Lebesgue measure for continuous treatment and counting measure for discrete treatment.

\begin{assumption}[Latent mean-IV model]
\label{ass:latent_iv_population}
The observations are iid and satisfy $Y=m_0(X,A_Y)+\epsilon$, with $Y,m_0(X,A_Y)\in L^2$ and $\mathbb E[\epsilon\mid V,W]=0$. The representation $A=\varphi(V)$ is sufficient for the structural mean in the sense stated above. The conditional law of $X$ given $(W,A)$ has density $k_0(x\mid w,a_X)$, so that its dependence on $a$ is through $a_X$. The causal interpretation of $m_0$ is inherited from the structural assumptions in Section~\ref{sec:problem_setup} and the stated sufficiency condition.
\end{assumption}

\begin{assumption}[Completeness on structural-mean differences]
\label{ass:latent_completeness}
For $P_A$-almost every $a$, define
\[
(T_a h)(w)=\int h(x)k_0(x\mid w,a_X)\,d\lambda_X(x).
\]
On the difference class $h(x)=m(x,a_Y)-m_0(x,a_Y)$ under consideration,
$(T_a h)(w)=0$ for $P_{W\mid A=a}$-almost every $w$ implies
$h(x)=0$ for $P_{X\mid A=a}$-almost every $x$.
\end{assumption}

\begin{lemma}[Exogeneity of fixed and learned covariate readouts]
\label{lem:latent_exogeneity}
For a fixed measurable map $A=\varphi(V)$, $\mathbb E[\epsilon\mid V,W]=0$ implies $\mathbb E[\epsilon\mid W,A]=0$.
More generally, let $\mathcal D$ contain training observations, fitted preprocessing, model selection and training randomness. For a new observation independent of $\mathcal D$, let $R$ be inference randomness independent of the joint tuple $(\mathcal D,V,W,\epsilon)$ and let $A=\varphi_{\mathcal D}(V,R)$. If $\mathbb E[\epsilon\mid V,W]=0$, then
\[
\mathbb E[\epsilon\mid W,A,\mathcal D]=0.
\]
The same conclusion applies to any measurable sub-block of $A$.
\end{lemma}
\begin{proof}
For the fixed map, $\sigma(W,A)\subseteq\sigma(W,V)$, so iterated expectations give the claim. For the learned map, independence implies
$\mathbb E[\epsilon\mid V,W,\mathcal D,R]=0$.
Since $\sigma(W,A,\mathcal D)\subseteq\sigma(V,W,\mathcal D,R)$, another conditional expectation proves the result. Sub-blocks generate smaller sigma-fields and inherit the same conclusion.
\end{proof}

\begin{remark}[Training states and evaluation readouts]
\label{rem:latent_exogeneity_scope}
The lemma permits training the network parameters with outcomes. It covers the numerical covariate-only MAP rule in Section~\ref{sec:model_inference}, the encoder, and posterior sampling with independent inference randomness: each uses only the new unit's covariates and fitted quantities. For a reused training observation, the fitted parameters may depend on that observation's error, so the new-unit argument does not apply directly. Appendix~\ref{app:latent_exogeneity} diagnoses associations in the fitted MAP readout on both training and held-out samples. The outcome-informed particles in Eq.~\eqref{eq:latent_update_objective} enter the training IV term and are re-inferred from covariates for final prediction. The lemma concerns conditional mean exogeneity of that prediction readout; information sufficiency and consistency of the paired outcome function are separate conditions.
\end{remark}

\begin{proposition}[Latent nonparametric IV moment]
\label{prop:latent_iv_moment}
Under Assumption~\ref{ass:latent_iv_population},
\begin{equation}
\mathbb E[Y\mid W=w,A=a]
=\int m_0(x,a_Y)k_0(x\mid w,a_X)\,d\lambda_X(x).
\label{eq:appendix_npiv_moment}
\end{equation}
Under Assumption~\ref{ass:latent_completeness}, $m_0$ is the unique solution in the stated class, up to equality for $P_A$-almost every $a$ and $P_{X\mid A=a}$-almost every $x$.
\end{proposition}
\begin{proof}
Lemma~\ref{lem:latent_exogeneity} gives $\mathbb E[\epsilon\mid W,A]=0$. Taking the conditional expectation of $Y=m_0(X,A_Y)+\epsilon$ and using the true first-stage law proves Eq.~\eqref{eq:appendix_npiv_moment}. If another $m$ satisfies the same equation, subtraction gives $T_a\{m(\cdot,a_Y)-m_0(\cdot,a_Y)\}=0$. Completeness on these differences proves uniqueness on the stated support.
\end{proof}

\subsection{Population Target of the Integrated EGM Outcome Component}

The integrated squared loss targets the IV mean restriction under a fixed sufficient representation and a correct first stage.

\begin{proposition}[Population target of the integrated EGM outcome component]
\label{prop:egm_integral_optimality}
Fix $A=\varphi(V)$ and suppose $Y=m_0(X,A_Y)+\epsilon$ with $\mathbb E[\epsilon\mid V,W]=0$. Assume $Y,m_0(X,A_Y)\in L^2$, all candidates satisfy $m(X,A_Y)\in L^2$, $m_0$ is in the candidate class, and the first-stage kernel in the objective equals the full conditional law $P(X\mid W,V)$ through $A_X$. For
\[
R(m)=\mathbb E\left[\left\{Y-\int m(x,A_Y)k_0(x\mid W,A_X)\,d\lambda_X(x)\right\}^2\right],
\]
the minimum is $\mathbb E[\operatorname{Var}(Y\mid W,V)]$. Under completeness on candidate mean differences, every minimizer equals $m_0$ on the identified support.
\end{proposition}
\begin{proof}
Put $T_m(W,V)=\int m(x,A_Y)k_0(x\mid W,A_X)\,d\lambda_X(x)$.
The conditional-mean restriction and correct first stage imply
$\mathbb E[Y\mid W,V]=T_{m_0}(W,V)$.
Conditional-mean orthogonality gives
\[
R(m)=\mathbb E[\operatorname{Var}(Y\mid W,V)]
+\|T_m-T_{m_0}\|_{L^2(P_{W,V})}^2.
\]
The second term vanishes at $m_0$. If it vanishes at another $m$, condition the resulting zero conditional moment on $(W,A)$ and apply completeness to $m-m_0$.
\end{proof}

\begin{remark}[Relation to EGM initialization]
Proposition~\ref{prop:egm_integral_optimality} characterizes the integrated outcome loss for a fixed sufficient representation and a correct first-stage kernel. EGM uses this loss together with its reconstruction, adversarial, and first-stage objectives. The final covariate-only MAP predictor is analyzed in Proposition~\ref{prop:final_map_selection} and Corollary~\ref{cor:quasi_posterior_consistency}.
\end{remark}

\subsection{Final Predictors and Conditional Consistency}

The following results concern the final outcome function composed with the covariate-only readout. Population optimality at a fixed representation and exogeneity of the final readout address distinct parts of the argument; a learning condition must connect the fitted composite predictor to the identified moment target.

\begin{remark}[A quantitative training-to-readout condition]
Let $g_E(x,v)=\mu_{\omega_E}(x,e_E(v)_Y)$ be an EGM composite predictor and let $g_F(x,v)=\mu_{\omega_F}(x,a_F(v)_Y)$ be the final readout. Define the true observed-covariate operator
$(T_0g)(w,v)=\mathbb E[g(X,V)\mid W=w,V=v]$.
Conditional expectation is an $L^2$ contraction, so
\[
\|T_0(g_F-g_0)\|_2\le\|T_0(g_E-g_0)\|_2+\|g_F-g_E\|_2.
\]
For outcome maps uniformly Lipschitz in their parameters and latent inputs on the relevant region, the last term is bounded by
\[
L_\omega\|\omega_F-\omega_E\|+L_z\|a_F-e_E\|_{L^2(P_V)}.
\]
This inequality isolates the transfer error; it does not assert that the reported fixed training schedule makes that error vanish. Exogeneity of $a_F(V)$ alone cannot control changes in its paired outcome map.
\end{remark}
\paragraph{The final selection criterion.}
We now state a result for the prediction functions actually compared by the
training-side selection rule, rather than for numerical latent coordinates.
For candidate $k$, write
\[
g_k(x,v)=\mu_{\widehat\omega_k}
 \bigl(x,\widehat z_{k,0}(v),\widehat z_{k,1}(v)\bigr),\qquad
\widehat K_k(dx\mid w,v)
=p_{\widehat\phi_k}
 \bigl(dx\mid w,\widehat z_{k,0}(v),\widehat z_{k,2}(v)\bigr).
\]
The covariate-only inference rule is assumed to define the same measurable function $v\mapsto\widehat z_k(v)$ in scoring and deployment, with the same fitted preprocessing, initialization and numerical inference settings. The results below concern this implemented prediction function. All training, preprocessing and model-selection inputs remain in the training sample. Assume $Y,g_0(X,V)\in L^2$ and define
\[
(T_0g)(w,v)=\int g(x,v)P_0(dx\mid w,v),\qquad
\mathcal B(g)=\|T_0(g-g_0)\|_{L^2(P_{W,V})}^2.
\]
The original mean-IV restriction gives the projection identity
\begin{equation}
R_0(g):=\mathbb E[(Y-T_0g(W,V))^2]
=C+\mathcal B(g),\qquad
C=\mathbb E[\operatorname{Var}(Y\mid W,V)].
\label{eq:final_readout_risk_identity}
\end{equation}

Let the training sample consist of $n$ iid observations, let $P_n$ be its empirical measure, and set $\widehat h_k(w,v)=\int g_k(x,v)\widehat K_k(dx\mid w,v)$. Conditional on the fitted candidates and training data, draw treatment values independently across observations and Monte Carlo replicates, using randomness not used to fit the candidates. Denote the resulting implemented squared-residual score by $S_k^{\rm MC}$ and set
\[
\nu_k=\frac1{nM}\sum_{i=1}^n
 \operatorname{Var}_{\widehat K_k(\cdot\mid W_i,V_i)}
       \{g_k(X,V_i)\}.
\]
Assume these conditional second moments and all displayed risks are finite.
The score then satisfies the exact identity
\begin{equation}
\mathbb E_{\rm MC}[S_k^{\rm MC}\mid\text{data and fitted candidates}]
=P_n(Y-\widehat h_k)^2+\nu_k.
\label{eq:selection_mc_variance_bias}
\end{equation}
For a deterministic class $\mathcal G_n\subset L^2(P_{X,V})$ containing
all possible fitted candidates, define
\begin{align*}
u_n&=\sup_{g\in\mathcal G_n}
 \bigl|P_n(Y-T_0g)^2-R_0(g)\bigr|,\\
e_n&=\max_{k\le K}
 \bigl|P_n(Y-\widehat h_k)^2-P_n(Y-T_0g_k)^2\bigr|,\\
t_n&=\max_{k\le K}
 \bigl|S_k^{\rm MC}-P_n(Y-\widehat h_k)^2-\nu_k\bigr|.
\end{align*}
These quantities separate empirical generalization, first-stage error at the final readout, and Monte Carlo fluctuation.

\begin{proposition}[Oracle bound for final-readout selection]
\label{prop:final_map_selection}
Under the preceding conditions, if
$\widehat k\in\arg\min_{k\le K}S_k^{\rm MC}$, then, pathwise,
\begin{equation}
\mathcal B(g_{\widehat k})
\le\min_{k\le K}\{\mathcal B(g_k)+\nu_k\}
 +2u_n+2e_n+2t_n.
\label{eq:final_map_oracle}
\end{equation}
\end{proposition}
\begin{proof}
Equation~\eqref{eq:final_readout_risk_identity} follows by expanding
$Y-T_0g=(Y-\mathbb E[Y\mid W,V])+T_0(g_0-g)$; the cross term has
zero expectation. Equation~\eqref{eq:selection_mc_variance_bias} follows
by taking the variance of the Monte Carlo mean. For every comparator $k$,
the definitions of $u_n,e_n,t_n$ and the selection rule give
\begin{align*}
R_0(g_{\widehat k})
&\le S_{\widehat k}^{\rm MC}-\nu_{\widehat k}+u_n+e_n+t_n\\
&\le S_k^{\rm MC}+u_n+e_n+t_n\\
&\le R_0(g_k)+\nu_k+2u_n+2e_n+2t_n.
\end{align*}
The second line uses $\nu_{\widehat k}\ge0$. Subtract $C$ and minimize
over $k$.
\end{proof}

\begin{assumption}[Final-candidate approximation]
\label{ass:final_candidate_approximation}
For nonnegative sequences $a_n$ and $b_n^{\rm MC}$, with probability
tending to one there is a fitted candidate $k_n^\dagger$ such that
\[
\|T_0(g_{k_n^\dagger}-g_0)\|_{L^2(P_{W,V})}\le a_n,
\qquad \nu_{k_n^\dagger}\le b_n^{\rm MC}.
\]
\end{assumption}

To distinguish unique identification from stable recovery, define the
restricted inverse modulus
\[
\mathfrak m_n(r)=\sup\left\{
\|g-g_0\|_{L^2(P_{X,V})}:
g\in\mathcal G_n\cup\{g_0\},\quad
\|T_0(g-g_0)\|_{L^2(P_{W,V})}\le r
\right\}.
\]

The modulus quantifies how IV-moment error translates into
structural-function error on the candidate class, following
ill-posedness measures used in NPIV
analyses~\citep{xu2021dfiv,dikkala2020minimax}.
For example, if a finite constant $\tau_n$ satisfies
\[
\|g-g_0\|_{L^2(P_{X,V})}
\le
\tau_n\|T_0(g-g_0)\|_{L^2(P_{W,V})}
\qquad\text{for all }g\in\mathcal G_n,
\]
then $\mathfrak m_n(r)\le\tau_n r$.
This gives a linear error bound of the form used by
\citet{xu2021dfiv}; the modulus formulation also accommodates
nonlinear error bounds.

Completeness alone does not make this modulus small uniformly over a
growing class. For a fixed $L^2$-compact class containing $g_0$,
continuity of $T_0$ and uniqueness of its moment solution imply
$\mathfrak m(r)\to0$ as $r\downarrow0$: otherwise compactness would
produce a distinct limit with the same conditional moments.

\begin{corollary}[Conditional consistency of the selected MAP predictor]
\label{cor:quasi_posterior_consistency}
Under Proposition~\ref{prop:final_map_selection} and
Assumption~\ref{ass:final_candidate_approximation}, let
\[
\Delta_n=a_n^2+b_n^{\rm MC}+2u_n+2e_n+2t_n.
\]
If $\mathfrak m_n(\sqrt{\Delta_n})\to_p0$, then
\begin{equation}
\|g_{\widehat k}-g_0\|_{L^2(P_{X,V})}\longrightarrow_p0.
\label{eq:final_map_conditional_consistency}
\end{equation}
\end{corollary}
\begin{proof}
On the event in Assumption~\ref{ass:final_candidate_approximation},
Eq.~\eqref{eq:final_map_oracle} gives
$\mathcal B(g_{\widehat k})\le\Delta_n$.
The definition of $\mathfrak m_n$ bounds the structural error by
$\mathfrak m_n(\sqrt{\Delta_n})$. The complement of that event has
probability tending to zero.
\end{proof}

\begin{remark}[Conditions and scope]
The uniform term $u_n$ handles reuse of training observations; no exact
same-sample latent-exogeneity claim is needed. A fixed uniformly bounded,
sup-norm totally bounded class with $\mathbb E Y^2<\infty$ satisfies the
required uniform law, whereas growing neural classes require additional
complexity control. For bounded candidate outputs $|g_k|\le B$, the
Monte Carlo bias obeys $\nu_k\le B^2/M$. Conditional on a sample with
$\max_i|Y_i|\le A$, bounded differences additionally give
\[
t_n\le B(A+B)\sqrt{\frac{8\log(2K/\delta)}{nM}}
\]
with conditional probability at least $1-\delta$. Thus increasing $n$
can reduce score fluctuation without removing the candidate-dependent
bias at fixed $M$. The corollary retains that term explicitly.

Assumption~\ref{ass:final_candidate_approximation} describes the approximation quality of the fitted candidates under the final covariate-only MAP readout. Corollary~\ref{cor:quasi_posterior_consistency} combines this condition with first-stage, empirical, and Monte Carlo error control and the restricted inverse modulus to establish $L^2(P_{X,V})$ consistency of the selected predictor.
\end{remark}
\subsection{A Complementary Observed-Data Comparison}
The candidate-approximation condition above can also be investigated through a reduced-form comparison after the original selection has been completed. This comparison is an optional diagnostic of the unchanged selected predictor, not a replacement selection rule or an experiment reported here. Let $g$ be that predictor and let $\widehat K$ be its fitted treatment kernel at the covariate-only readout. Put
\[
m_{\mathrm{RF}}(w,v)=\mathbb E[Y\mid W=w,V=v],\quad
h_0=T_0g,\quad \widehat h=\int g(x,v)\widehat K(dx\mid w,v).
\]
Let $\widehat m_{\mathrm{RF}}$ be a benchmark fitted using observed
variables, with $a_n^{\mathrm{RF}}=\|\widehat m_{\mathrm{RF}}-m_{\mathrm{RF}}\|_2$.
After fixing both the selected model and this benchmark, generate fresh
draws $X_{im}$ from $\widehat K(\cdot\mid W_i,V_i)$, conditionally
independently over $(i,m)$. For $M\ge2$, write
$G_{im}=g(X_{im},V_i)$, $\overline G_i=M^{-1}\sum_mG_{im}$, and define
\[
\mathcal U_n=\frac1n\sum_i
\left\{(Y_i-\overline G_i)^2-\frac{s_{G,i}^2}{M}\right\},
\quad
s_{G,i}^2=\frac1{M-1}\sum_m(G_{im}-\overline G_i)^2,
\]
\[
\widehat D_n=\mathcal U_n-P_n(Y-\widehat m_{\mathrm{RF}})^2.
\]
The diagnostic gap $\widehat D_n$ uses no structural truth. For deterministic classes $\mathcal G_n,\mathcal H_n$ containing all possible selected predictors and fitted benchmarks, respectively, define
\begin{align*}
u_n^G&=\sup_{g'\in\mathcal G_n}
 |(P_n-P)(Y-T_0g')^2|,\\
u_n^{\mathrm{RF}}&=\sup_{m\in\mathcal H_n}|(P_n-P)(Y-m)^2|.
\end{align*}
Put
\[
e_n^{\mathrm{stage}}=
|P_n(Y-\widehat h)^2-P_n(Y-h_0)^2|,\qquad
\zeta_n=\mathcal U_n-P_n(Y-\widehat h)^2.
\]

\begin{proposition}[Reduced-form comparison bound]
\label{prop:observable_moment_comparison}
Under the mean-IV restriction, the stated uniform bounds, square-integrable
risks and finite conditional second moments of the diagnostic draws,
\begin{equation}
\left|\mathcal B(g)-\widehat D_n-(a_n^{\mathrm{RF}})^2\right|
\le u_n^G+u_n^{\mathrm{RF}}+e_n^{\mathrm{stage}}+|\zeta_n|.
\label{eq:observable_moment_comparison}
\end{equation}
Consequently, with $[u]_+=\max(u,0)$,
\[
\mathcal B(g)\le[\widehat D_n]_++(a_n^{\mathrm{RF}})^2+
u_n^G+u_n^{\mathrm{RF}}+e_n^{\mathrm{stage}}+|\zeta_n|.
\]
Applying the restricted inverse modulus to the square root of this upper bound gives a sufficient condition for structural consistency of this already selected predictor, without an assumption about another good candidate.
\end{proposition}
\begin{proof}
Conditional-mean orthogonality gives
\[
\mathcal B(g)=R_0(g)
-\mathbb E[(Y-\widehat m_{\mathrm{RF}}(W,V))^2]+(a_n^{\mathrm{RF}})^2.
\]
Conditional on the fitted quantities and data, the rowwise diagnostic equals
\[
\frac1{M(M-1)}\sum_{m\ne l}(Y_i-G_{im})(Y_i-G_{il}),
\]
whose expectation is $(Y_i-\widehat h(W_i,V_i))^2$ by independence. Insert the two empirical risks and this conditional expectation into the preceding population identity. The triangle inequality gives Eq.~\eqref{eq:observable_moment_comparison}; replacing the observed gap by its positive part can only increase the upper bound.
\end{proof}

Fresh diagnostic draws after selection are essential for the displayed
conditional expectation. If $v_i=\operatorname{Var}(G_{im}\mid
\text{data and fitted quantities})$ and $a_i=Y_i-\widehat h(W_i,V_i)$,
the conditional diagnostic variance is exactly
\[
\operatorname{Var}(\mathcal U_n\mid\text{data and fitted quantities})
=\frac1{n^2}\sum_i\left\{
\frac{4a_i^2v_i}{M}+\frac{2v_i^2}{M(M-1)}\right\}.
\]
To obtain this formula, center each draw and expand the products; distinct centered pairs are uncorrelated, including pairs sharing one index. With suitable moment control, the diagnostic fluctuation vanishes even at fixed $M\ge2$. This does not remove the variance term from the original selection score, which is unchanged.

The observable gap alone is not a complete confidence certificate: benchmark consistency, first-stage accuracy, uniform generalization and inverse stability still require justification. A poor benchmark can make the gap misleadingly small, which is why $(a_n^{\mathrm{RF}})^2$ is retained. Nor does the current algorithm automatically make the diagnostic gap tend to zero. Using the same observations for fitting and this diagnostic requires the uniform bounds above; fixing fitted parameters afterwards does not establish sample independence. This comparison is not used here to reselect candidates or tune the estimator.

\FloatBarrier
\section{Empirical Check of Latent-State Exogeneity}
\label{app:latent_exogeneity}

Lemma~\ref{lem:latent_exogeneity} establishes conditional mean exogeneity of the covariate-only readout for independent new units. This appendix checks the argument empirically on the BGM-IV models of the demand design benchmark of Section~\ref{sec:demand_design}, whose unobserved confounder is written explicitly and can be replayed exactly.

\paragraph{Design.}
In the demand design the structural error is $\epsilon=\rho U+\sqrt{1-\rho^2}\,\xi$, where $U$ is the unobserved demand shock that also enters the price equation and $\xi$ is an independent standard normal disturbance. Because the price equation is $P=25+(C+3)\psi(T)+U$, the shock can be recovered exactly in the simulation as $U=P-25-(C+3)\psi(T)$ from the observed $(P,C,T)$ and the known seasonal function $\psi$, so the only part of $Y$ that cannot be inferred from pre-outcome observables is $\sqrt{1-\rho^2}\,\xi$. The disturbance $\xi$ is independent of $(X,V,W)$ and enters the data only through $Y$, an association between a latent coordinate and $\xi$ beyond sampling variability would therefore indicate that the latent state has absorbed outcome information. We correlate the latent states with both $\epsilon$ and $\xi$, and separately with $U$ to check whether the latent representation encodes the unobserved confounder itself. We report the strongest-confounding setting, $\rho=0.9$, where $U$ accounts for $81\%$ of the variance of $\epsilon$ and deconfounding is hardest.

\paragraph{Protocol.}
The models are the twenty repetitions of Table~\ref{tab:demand_mse} at $\rho=0.9$ for training sample sizes $n=5{,}000$ and $n=10{,}000$, each selected among ten candidates by the criterion of Eq.~\eqref{eq:selection_criterion}. Every model was restored from its archived checkpoint, the restored models reproduce their reported structural MSE (largest absolute difference $1.8\times10^{-5}$) on training data that is bit-identical to the archived run. Latent states are inferred by the covariate-only MAP of Eq.~\eqref{eq:test_time_map}, exactly as at evaluation, on two samples: the training sample of each model, and one held-out sample of $5{,}000$ units generated from an independent seed and shared by all models in both columns of Table~\ref{tab:latent_exogeneity}. For each model we compute the Pearson correlation between each of the eight latent coordinates and each of $\epsilon$ and $\xi$, giving $20\times8\times2=320$ coordinate-level tests per column. A correlation is called \emph{outside the null band} when $|r|>1.96/\sqrt{m}$, with $m$ the number of units in the sample on which it is computed ($\pm0.0277$ on the held-out sample and on the $n=5{,}000$ training sample, $\pm0.0196$ on the $n=10{,}000$ training sample). If a latent coordinate were unrelated to the target, about $5\%$ of these correlations would fall outside the band by chance, which is the nominal level of the test. Therefore, we compare the observed fraction outside the band (\emph{Outside the null band} in Table~\ref{tab:latent_exogeneity}) with this $5\%$ reference, following the standard residual diagnostic for white noise~\citep{hyndman2014forecasting}. Multiplicity is handled by Holm's procedure~\citep{holm1979simple} within families defined by one model and one target ($\epsilon$ or $\xi$) over its eight coordinates, i.e.\ 40 families per column. We count the tests that are outside the band and remain significant after the correction.

\paragraph{Controls.}
Three controls calibrate the tests. Negative controls replace the latent coordinate by the instrument $W$, the time covariate $T$, the customer group $S$, and an independent $\mathcal N(0,1)$ variable, and apply the same correlation tests to each of them against $\epsilon$ and $\xi$ on the training sample ($160$ tests per column). Their true correlation is zero, so their rejection rate shows the behavior of the procedure under the null~\citep{WOS:000277071500016}. Positive controls inject a known correlation with $\xi$ into the first coordinate of $\mathbf z_0$ on the training sample, at $r=0.03$ and $0.05$ for $n=5{,}000$ ($1.08\times$ and $1.80\times$ the training null band) and $r=0.021$ and $0.035$ for $n=10{,}000$ ($1.07\times$ and $1.79\times$). An injection counts as detected when the $F$ test of the $\mathbf z_0$ block against $\xi$ has $p<0.05$. Finally, $\mathrm{corr}(\epsilon,U)$ is the unobserved confounding present in the data and must be detected: its mean over the twenty models is $0.90$ (range $0.896$--$0.906$ at $n=5{,}000$ and $0.896$--$0.904$ at $n=10{,}000$).

\begin{table}[!ht]
\centering
\caption{Latent-state exogeneity check on the demand design at $\rho=0.9$. Columns are the training sample sizes of the twenty BGM-IV models ($\gamma=0.01$, ten candidates with training-set selection). Each percentage has denominator $320=20$ models $\times$ $8$ latent coordinates $\times$ $\{\epsilon,\xi\}$, with latent states inferred by the covariate-only MAP. The held-out block uses one sample of $5{,}000$ units shared by both columns; the null band is $\pm1.96/\sqrt{m}$ for a sample of $m$ units ($\pm0.0277$ held-out and for the $n=5{,}000$ training sample, $\pm0.0196$ for the $n=10{,}000$ training sample), and ``largest $|r|$ / null band'' is relative to the band of the sample in question. Holm correction is applied within each model and target over its eight coordinates (40 families per column). Negative controls are $W$, $T$, $S$, and an independent $\mathcal N(0,1)$ variable, each tested for correlation with $\epsilon$ and $\xi$ ($160$ tests per column). Injected signals are correlations with $\xi$ added to the first coordinate of $\mathbf z_0$ on the training sample ($r=0.03/0.05$ at $n=5{,}000$, $r=0.021/0.035$ at $n=10{,}000$), counted as detected when the $\mathbf z_0$-block $F$ test has $p<0.05$. $\mathrm{corr}(\epsilon,U)$ is the mean over the twenty models.}
\label{tab:latent_exogeneity}
\setlength{\tabcolsep}{6pt}
\begin{tabular}{@{}lcc@{}}
\toprule
 & $n=5{,}000$ & $n=10{,}000$ \\
\midrule
\multicolumn{3}{@{}l}{\textit{Held-out sample (5{,}000 units, shared by both columns)}} \\
\quad Outside the null band (about 5\% if unrelated) & 1.9\% (6/320) & 0.3\% (1/320) \\
\quad Significant after within-family Holm correction & 0/320 & 0/320 \\
\quad Largest $|r|$ / null band & 1.25$\times$ & 1.26$\times$ \\
\quad Largest $|\mathrm{corr}(z, U)|$ & 0.035 & 0.028 \\
\midrule
\multicolumn{3}{@{}l}{\textit{Training sample}} \\
\quad Outside the null band (about 5\% if unrelated) & 4.1\% (13/320) & 4.1\% (13/320) \\
\quad Significant after within-family Holm correction & 0/320 & 1/320 \\
\quad Largest $|r|$ / null band & 1.26$\times$ & 1.87$\times$ \\
\quad Largest $|\mathrm{corr}(z, U)|$ & 0.036 & 0.027 \\
\midrule
\multicolumn{3}{@{}l}{\textit{Controls (training sample)}} \\
\quad Negative controls outside the null band & 5.0\% (8/160) & 6.9\% (11/160) \\
\quad Injected signal at $\approx$1.1$\times$ null band detected & 3/20 & 6/20 \\
\quad Injected signal at $\approx$1.8$\times$ null band detected & 20/20 & 20/20 \\
\quad $\mathrm{corr}(\epsilon, U)$ (/ null band) & 0.90 (32--33$\times$) & 0.90 (46$\times$) \\
\bottomrule
\end{tabular}
\end{table}

\paragraph{Results.}
Table~\ref{tab:latent_exogeneity} summarizes the outcome. On the held-out sample, $1.9\%$ ($n=5{,}000$) and $0.3\%$ ($n=10{,}000$) of the $320$ coordinate-level correlations fall outside the null band, none survives the Holm correction, and the largest correlation is $1.26$ times the band. On the training sample, where the network weights were fitted with the outcomes of the very same subjects, the rejection rates are $4.1\%$ in both columns, below the nominal $5\%$ and within the range of the negative controls ($5.0\%$ and $6.9\%$). Only one of the $320$ tests at $n=10{,}000$ remains significant after Holm correction, whereas under the null hypothesis at most two of the 40 families per column are expected to contain a rejection. The positive controls confirm that the tests have adequate statistical power at the relevant effect sizes, the injected correlations at about $1.8$ times the band are detected in all twenty models, and $\mathrm{corr}(\epsilon,U)$ is detected at $32$ to $46$ times the band. Injections at about $1.1$ times the band, which lie at the detection threshold of a $5\%$-level test, are detected in $3/20$ and $6/20$ models, as expected for a signal of that size. The check therefore resolves associations of about twice the null band and larger, which is why we describe the result as no detectable association. Finally, the latent states carry no detectable information about the unobserved confounder: the largest $|\mathrm{corr}(z,U)|$ is $0.036$, and the linear $R^2$ of $U$ on the $\mathbf z_0$ block is at most $0.002$. The held-out results are consistent with Lemma~\ref{lem:latent_exogeneity}; the training-sample results provide a diagnostic of outcome-related associations in the re-inferred prediction states.

\FloatBarrier
\section{Sensitivity Analysis}
\label{app:gamma_sensitivity}

\begin{figure}[!ht]
  \centering
  \includegraphics[width=\linewidth]{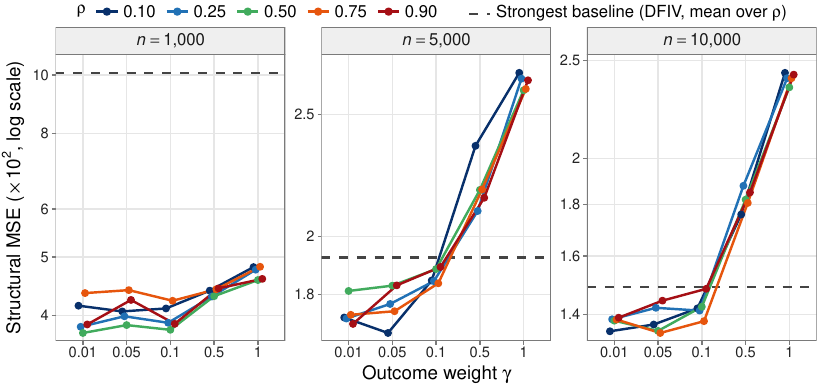}
  \caption{Sensitivity of BGM-IV to the tempering exponent $\gamma$ on the demand-design benchmark. Panels correspond to training sample sizes $n=1{,}000$, $5{,}000$, and $10{,}000$; each line is one confounding level $\rho$, and each point is the mean structural MSE over 20 repetitions (standard deviations are given in Table~\ref{tab:demand_gamma}). The five values of $\gamma$ are placed at equal horizontal spacing and the vertical axis is logarithmic, in units of $10^{2}$. The dashed line marks the strongest baseline in each panel (DFIV), averaged over $\rho$.}
  \label{fig:demand_gamma}
\end{figure}

The exponent $\gamma$ in Eq.~\eqref{eq:iv_quasi_posterior} sets the weight of the IV-integrated pseudo-likelihood in the latent update of Eq.~\eqref{eq:latent_update_objective}, it does not enter the objectives of the generators. All experiments in Section~\ref{sec:experiments} use $\gamma=0.01$. Figure~\ref{fig:demand_gamma} and Table~\ref{tab:demand_gamma} examine the sensitivity of BGM-IV to this choice on the demand-design benchmark over the full grid of sample sizes and confounding levels, for $\gamma\in\{0.01,0.05,0.1,0.5,1\}$, where $\gamma=1$ uses the pseudo-likelihood without tempering. Only $\gamma$ differs between the five settings, within each repetition they share the training data and the initialization seeds, and each is trained with the multi-start protocol of Appendix~\ref{app:baseline_selection}, so the comparison is paired at the level of repetitions.

The three smallest values, $\gamma\in\{0.01,0.05,0.1\}$, give structural MSEs that differ by less than the repetition-to-repetition variability in every cell, with no consistent ordering across sample sizes or confounding levels. Larger weights degrade the estimates monotonically, and the degradation grows with the sample size, relative to $\gamma=0.01$, the mean MSE at $\gamma=1$ is higher by a factor of $1.18$ at $n=1{,}000$, $1.52$ at $n=5{,}000$, and $1.74$ at $n=10{,}000$, and $\gamma=0.01$ has the lower MSE in $270$ of the $300$ paired repetitions ($196$ of $200$ at $n\geq5{,}000$). The comparison with the baselines of Table~\ref{tab:demand_mse} follows the same pattern, $\gamma=0.01$ attains the lowest mean MSE of all methods in every cell and $\gamma\in\{0.05,0.1\}$ in $14$ of $15$, whereas $\gamma\in\{0.5,1\}$ fall behind DFIV in all ten cells with $n\geq5{,}000$. Giving the pseudo-likelihood term a weight close to that of a proper likelihood therefore harms the structural estimate, whereas a small positive weight leaves the results in a flat region that contains $\gamma=0.01$. The selection criterion of Eq.~\eqref{eq:selection_criterion}, which uses only training data, gives the same ordering: in every cell, its value for the selected candidate, averaged over the twenty repetitions, is higher at $\gamma=0.5$ and at $\gamma=1$ than at each of $\gamma\in\{0.01,0.05,0.1\}$. The ordering among the three small values changes from cell to cell, so the criterion supports a small $\gamma$ but does not select one value within that range. A small $\gamma$ also limits the weight of the outcome in the latent update during training, consistent with the exogeneity check of Appendix~\ref{app:latent_exogeneity}.

\begin{table}[!ht]
\centering
\caption{Sensitivity of BGM-IV to the tempering exponent $\gamma$ on the demand-design benchmark. Entries are structural MSE divided by $10^{2}$, mean $\pm$ sample standard deviation over 20 repetitions, grouped by training sample size $n$. Only $\gamma$ differs between columns, within each repetition the five settings use the same training data and initialization seeds, and each is selected among ten candidates by the training-set criterion. The lowest mean in each row is in bold.}
\label{tab:demand_gamma}
\begin{tabular}{@{}lccccc@{}}
\toprule
$\rho$ & $\gamma=0.01$ & $\gamma=0.05$ & $\gamma=0.1$ & $\gamma=0.5$ & $\gamma=1$ \\
\midrule
\multicolumn{6}{@{}l}{$n=1{,}000$} \\
\addlinespace[1pt]
0.1 & 4.15 $\pm$ 0.91 & \textbf{4.06 $\pm$ 0.72} & 4.11 $\pm$ 0.80 & 4.40 $\pm$ 0.79 & 4.81 $\pm$ 0.74 \\
0.25 & \textbf{3.83 $\pm$ 0.69} & 3.99 $\pm$ 0.79 & 3.89 $\pm$ 0.74 & 4.30 $\pm$ 0.84 & 4.76 $\pm$ 1.07 \\
0.5 & \textbf{3.74 $\pm$ 0.47} & 3.86 $\pm$ 0.52 & 3.79 $\pm$ 0.52 & 4.31 $\pm$ 0.84 & 4.58 $\pm$ 0.83 \\
0.75 & 4.36 $\pm$ 0.70 & 4.41 $\pm$ 1.06 & \textbf{4.23 $\pm$ 0.78} & 4.42 $\pm$ 0.74 & 4.82 $\pm$ 0.89 \\
0.9 & \textbf{3.87 $\pm$ 0.69} & 4.25 $\pm$ 0.79 & 3.88 $\pm$ 0.55 & 4.44 $\pm$ 1.08 & 4.60 $\pm$ 1.13 \\
\midrule
\multicolumn{6}{@{}l}{$n=5{,}000$} \\
\addlinespace[1pt]
0.1 & 1.72 $\pm$ 0.25 & \textbf{1.68 $\pm$ 0.19} & 1.85 $\pm$ 0.24 & 2.36 $\pm$ 0.55 & 2.70 $\pm$ 0.59 \\
0.25 & \textbf{1.72 $\pm$ 0.37} & 1.77 $\pm$ 0.40 & 1.84 $\pm$ 0.43 & 2.10 $\pm$ 0.28 & 2.67 $\pm$ 0.64 \\
0.5 & \textbf{1.81 $\pm$ 0.38} & 1.83 $\pm$ 0.36 & 1.88 $\pm$ 0.31 & 2.18 $\pm$ 0.32 & 2.61 $\pm$ 0.46 \\
0.75 & \textbf{1.73 $\pm$ 0.39} & 1.74 $\pm$ 0.36 & 1.84 $\pm$ 0.38 & 2.18 $\pm$ 0.39 & 2.62 $\pm$ 0.47 \\
0.9 & \textbf{1.70 $\pm$ 0.22} & 1.83 $\pm$ 0.30 & 1.89 $\pm$ 0.32 & 2.15 $\pm$ 0.38 & 2.66 $\pm$ 0.52 \\
\midrule
\multicolumn{6}{@{}l}{$n=10{,}000$} \\
\addlinespace[1pt]
0.1 & \textbf{1.35 $\pm$ 0.18} & 1.37 $\pm$ 0.21 & 1.42 $\pm$ 0.23 & 1.76 $\pm$ 0.21 & 2.43 $\pm$ 0.42 \\
0.25 & \textbf{1.38 $\pm$ 0.25} & 1.42 $\pm$ 0.27 & 1.41 $\pm$ 0.24 & 1.88 $\pm$ 0.37 & 2.40 $\pm$ 0.36 \\
0.5 & 1.38 $\pm$ 0.26 & \textbf{1.35 $\pm$ 0.31} & 1.42 $\pm$ 0.25 & 1.82 $\pm$ 0.28 & 2.35 $\pm$ 0.41 \\
0.75 & 1.38 $\pm$ 0.25 & \textbf{1.34 $\pm$ 0.21} & 1.38 $\pm$ 0.34 & 1.81 $\pm$ 0.25 & 2.40 $\pm$ 0.38 \\
0.9 & \textbf{1.39 $\pm$ 0.31} & 1.44 $\pm$ 0.37 & 1.48 $\pm$ 0.34 & 1.85 $\pm$ 0.53 & 2.42 $\pm$ 0.64 \\
\bottomrule
\end{tabular}
\end{table}

\FloatBarrier
\section{Detailed Training and Evaluation Algorithm}
\label{app:detailed_algorithm}

Algorithms~\ref{alg:egm_appendix} and~\ref{alg:bgm_appendix} give the EGM initialization and the stochastic optimization training of Section~\ref{sec:training} for one candidate; Algorithm~\ref{alg:causalbgm_iv_appendix} runs them for $K$ candidates and applies the selection rule of Eq.~\eqref{eq:selection_criterion}, Algorithm~\ref{alg:inference_appendix} gives the structural prediction and uncertainty quantification of Section~\ref{sec:model_inference}. Hyperparameter values are listed in Appendix~\ref{app:bgm_iv_hyperparameters}.

\paragraph{IV-aware EGM initialization.}
The IV-aware EGM initialization provides the starting values of the generators and the latent states. It trains an encoder $e_\psi$ from the covariates to the latent space jointly with the three generators $(G,H,F)$ by minimizing
\begin{equation}
\mathcal L_{\mathrm{EGM}}(\psi,\theta,\phi,\omega)
=
\mathcal L_{\mathrm{adv}}+\mathcal L_{\mathrm{rec}}+\mathcal L_{x}+\mathcal L_{y},
\label{eq:egm_objective}
\end{equation}
where all terms are evaluated at the encoded state $\tilde{\mathbf z}=e_\psi(\mathbf v)$. The adversarial term $\mathcal L_{\mathrm{adv}}$ matches the distribution of $e_\psi(V)$ to the prior $p_Z$ through a discriminator trained with a gradient penalty, and the reconstruction term $\mathcal L_{\mathrm{rec}}$ penalizes the error of reconstructing $\mathbf v$ from $\tilde{\mathbf z}$ and of recovering a prior draw $\mathbf z$ through the generator--encoder cycle, as in EGM~\citep{liu2024encoding}. The remaining two terms make the initialization IV-aware. $\mathcal L_x$ is the first-stage squared error $\{x-\mu_\phi(w,\tilde{\mathbf z}_0,\tilde{\mathbf z}_2)\}^2$, and $\mathcal L_y$ is the integrated outcome moment
\begin{equation}
\Big\{y-\int \mu_\omega(x,\tilde{\mathbf z}_0,\tilde{\mathbf z}_1)\,p_\phi(x\mid w,\tilde{\mathbf z}_0,\tilde{\mathbf z}_2)\,dx\Big\}^2,
\label{eq:egm_outcome_moment}
\end{equation}
which applies the integral of Eq.~\eqref{eq:iv_conditional_mean} to the outcome mean rather than to the outcome density. Under the conditions of Proposition~\ref{prop:egm_integral_optimality}, the population minimizer of Eq.~\eqref{eq:egm_outcome_moment} is the structural mean, so the outcome component of the initialization targets the structural function rather than the regression of $Y$ on the observed treatment.

For a scalar treatment with a Gaussian first stage, the integral in Eq.~\eqref{eq:egm_outcome_moment} is one-dimensional and its integrand is smooth, and we evaluate it by Gauss--Hermite quadrature with nodes $t_q$ and normalized weights $\lambda_q$,
\[
\int \mu_\omega(x,\tilde{\mathbf z}_0,\tilde{\mathbf z}_1)\,p_\phi(x\mid w,\tilde{\mathbf z}_0,\tilde{\mathbf z}_2)\,dx
\approx
\sum_{q=1}^{Q}\lambda_q\,\mu_\omega\big(\mu_\phi(w,\tilde{\mathbf z}_0,\tilde{\mathbf z}_2)+\sqrt2\,\tilde\sigma\,t_q,\ \tilde{\mathbf z}_0,\tilde{\mathbf z}_1\big),
\]
where $\tilde\sigma$ is a running estimate of the first-stage residual scale. The stochastic optimization training uses Monte Carlo instead, as in Eq.~\eqref{eq:mc_iv_likelihood}, because the outcome density can be sharply peaked in $x$ when $\sigma_\omega$ is small. For a binary treatment the integral is an exact two-point sum over the treatment values weighted by $p_\phi$. At the end of the initialization the latent states are set to $\mathbf z^{(i)}\leftarrow e_\psi(\mathbf v^{(i)})$, and the encoder is not updated afterwards. The Algorithm~\ref{alg:egm_appendix} gives the complete procedure.

\begin{algorithm}[!ht]
\caption{IV-aware EGM initialization for one candidate}
\label{alg:egm_appendix}
\begin{algorithmic}[1]
\REQUIRE Training data $\mathcal D=\{(x^{(i)},y^{(i)},\mathbf v^{(i)},w^{(i)})\}_{i=1}^n$; prior $p(\mathbf z)$; initialization seed $s$; EGM iterations $T_{\mathrm{EGM}}$ with $n_d$ discriminator steps per generator step; Gauss--Hermite nodes and weights $\{(t_q,\lambda_q)\}_{q=1}^{Q}$; step size $\alpha_{\mathrm{EGM}}$.
\ENSURE Initialized generators $(\theta^{(0)},\phi^{(0)},\omega^{(0)})$ and latent states $\{\mathbf z^{(i,0)}\}_{i=1}^n$.
\STATE Initialize the encoder $e_\psi$, the discriminator $D_\kappa$, and the generators $(\theta,\phi,\omega)$ from seed $s$.
\FOR{$t=1,\ldots,T_{\mathrm{EGM}}$}
    \FOR{$n_d$ steps}
    \STATE Sample a mini-batch $\mathcal B\subset\{1,\ldots,n\}$ and prior draws $\mathbf z^{(i)}\sim p(\mathbf z)$, $i\in\mathcal B$, and set $\tilde{\mathbf z}^{(i)}\leftarrow e_\psi(\mathbf v^{(i)})$.
    \STATE Update $\kappa$ by gradient ascent on $|\mathcal B|^{-1}\sum_{i\in\mathcal B}\{D_\kappa(\mathbf z^{(i)})-D_\kappa(\tilde{\mathbf z}^{(i)})\}$ with a gradient penalty (WGAN-GP).
  \ENDFOR
  \STATE Sample a new mini-batch $\mathcal B\subset\{1,\ldots,n\}$ and new prior draws $\mathbf z^{(i)}\sim p(\mathbf z)$, $i\in\mathcal B$.
  \STATE Set $\tilde{\mathbf z}^{(i)}\leftarrow e_\psi(\mathbf v^{(i)})$ and partition $\tilde{\mathbf z}^{(i)}=(\tilde{\mathbf z}_0^{(i)},\tilde{\mathbf z}_1^{(i)},\tilde{\mathbf z}_2^{(i)},\tilde{\mathbf z}_3^{(i)})$ for $i\in\mathcal B$.
  \STATE Set $\mathcal L_{\mathrm{adv}}\leftarrow -|\mathcal B|^{-1}\sum_{i\in\mathcal B}D_\kappa(\tilde{\mathbf z}^{(i)})$.
  \STATE Set $\mathcal L_{\mathrm{rec}}\leftarrow |\mathcal B|^{-1}\sum_{i\in\mathcal B}\big[\|\mathbf v^{(i)}-\boldsymbol\mu_\theta(\tilde{\mathbf z}^{(i)})\|^2+\|\mathbf z^{(i)}-e_\psi(\boldsymbol\mu_\theta(\mathbf z^{(i)}))\|^2\big]$.
  \STATE Set $\mathcal L_x\leftarrow |\mathcal B|^{-1}\sum_{i\in\mathcal B}\{x^{(i)}-\mu_\phi(w^{(i)},\tilde{\mathbf z}_0^{(i)},\tilde{\mathbf z}_2^{(i)})\}^2$.
  \FOR{each $i\in\mathcal B$}
    \STATE Compute the quadrature estimate of the integrated outcome mean $\int\mu_\omega(x,\tilde{\mathbf z}_0^{(i)},\tilde{\mathbf z}_1^{(i)})\,p_\phi(x\mid w^{(i)},\tilde{\mathbf z}_0^{(i)},\tilde{\mathbf z}_2^{(i)})\,dx$:
    \STATE $\bar\mu^{(i)}\leftarrow\sum_{q=1}^{Q}\lambda_q\,\mu_\omega\big(\mu_\phi(w^{(i)},\tilde{\mathbf z}_0^{(i)},\tilde{\mathbf z}_2^{(i)})+\sqrt2\,\tilde\sigma\,t_q,\ \tilde{\mathbf z}_0^{(i)},\tilde{\mathbf z}_1^{(i)}\big)$, where $\tilde\sigma$ is the running first-stage residual scale.
  \ENDFOR
  \STATE Set $\mathcal L_y\leftarrow |\mathcal B|^{-1}\sum_{i\in\mathcal B}\{y^{(i)}-\bar\mu^{(i)}\}^2$.
  \STATE Update $(\psi,\theta,\phi,\omega)$ by one gradient step of size $\alpha_{\mathrm{EGM}}$ on $\mathcal L_{\mathrm{adv}}+\mathcal L_{\mathrm{rec}}+\mathcal L_x+\mathcal L_y$.
\ENDFOR
\STATE Set $\mathbf z^{(i,0)}\leftarrow e_\psi(\mathbf v^{(i)})$ for $i=1,\ldots,n$ and $(\theta^{(0)},\phi^{(0)},\omega^{(0)})\leftarrow(\theta,\phi,\omega)$.
\STATE Return $(\theta^{(0)},\phi^{(0)},\omega^{(0)})$ and $\{\mathbf z^{(i,0)}\}_{i=1}^n$.
\end{algorithmic}
\end{algorithm}

\begin{algorithm}[!ht]
\caption{Stochastic optimization training for one candidate}
\label{alg:bgm_appendix}
\begin{algorithmic}[1]
\REQUIRE Training data $\mathcal D$; prior $p(\mathbf z)$; initialized generators $(\theta^{(0)},\phi^{(0)},\omega^{(0)})$ and latent states $\{\mathbf z^{(i,0)}\}_{i=1}^n$ from Algorithm~\ref{alg:egm_appendix}; BGM epochs $T_{\mathrm{BGM}}$; Monte Carlo size $M$; step sizes $\alpha_\theta,\alpha_\phi,\alpha_\omega,\beta_z$; tempering exponent $\gamma$.
\ENSURE Trained generators $(\theta,\phi,\omega)$.
\STATE Set $\tau\leftarrow0$.
\FOR{epoch $=1,\ldots,T_{\mathrm{BGM}}$}
  \FOR{each mini-batch $\mathcal B\subset\{1,\ldots,n\}$}
    \STATE Partition $\mathbf z^{(i,\tau)}=(\mathbf z_0^{(i,\tau)},\mathbf z_1^{(i,\tau)},\mathbf z_2^{(i,\tau)},\mathbf z_3^{(i,\tau)})$ for $i\in\mathcal B$.
    \STATE Set $\mathcal L_V(\theta)\leftarrow -|\mathcal B|^{-1}\sum_{i\in\mathcal B}\log p_\theta(\mathbf v^{(i)}\mid\mathbf z^{(i,\tau)})$.
    \STATE Update $\theta^{(\tau+1)}\leftarrow\theta^{(\tau)}-\alpha_\theta\nabla_\theta\mathcal L_V(\theta^{(\tau)})$.
    \STATE Set $\mathcal L_X(\phi)\leftarrow -|\mathcal B|^{-1}\sum_{i\in\mathcal B}\log p_\phi(x^{(i)}\mid w^{(i)},\mathbf z_0^{(i,\tau)},\mathbf z_2^{(i,\tau)})$.
    \STATE Update $\phi^{(\tau+1)}\leftarrow\phi^{(\tau)}-\alpha_\phi\nabla_\phi\mathcal L_X(\phi^{(\tau)})$.
    \FOR{each $i\in\mathcal B$}
      \IF{$X$ is continuous}
        \STATE Draw $x^{(i,m)}\sim p_{\phi^{(\tau+1)}}(x\mid w^{(i)},\mathbf z_0^{(i,\tau)},\mathbf z_2^{(i,\tau)})$, $m=1,\ldots,M$.
        \STATE Evaluate $\log p^{(i,m)}\leftarrow\log p_{\omega^{(\tau)}}(y^{(i)}\mid x^{(i,m)},\mathbf z_0^{(i,\tau)},\mathbf z_1^{(i,\tau)})$.
        \STATE Set $c^{(i)}\leftarrow\max_m\log p^{(i,m)}$ and
        $\log\widehat L_{\mathrm{IV}}^{(i)}\leftarrow c^{(i)}+\log\{\sum_{m=1}^{M}\exp(\log p^{(i,m)}-c^{(i)})\}-\log M$, which evaluates Eq.~\eqref{eq:mc_iv_likelihood} using the standard log-sum-exp stabilization.
      \ELSE
        \STATE Compute $\log\widehat L_{\mathrm{IV}}^{(i)}$ by evaluating Eq.~\eqref{eq:iv_integrated_likelihood} as an exact two-point mixture over $x\in\{0,1\}$.
        \STATE The mixture uses weights from $p_{\phi^{(\tau+1)}}(x\mid w^{(i)},\mathbf z_0^{(i,\tau)},\mathbf z_2^{(i,\tau)})$ and component densities $p_{\omega^{(\tau)}}(y^{(i)}\mid x,\mathbf z_0^{(i,\tau)},\mathbf z_1^{(i,\tau)})$.
      \ENDIF
    \ENDFOR
    \STATE Set $\mathcal L_Y(\omega)\leftarrow -|\mathcal B|^{-1}\sum_{i\in\mathcal B}\log\widehat L_{\mathrm{IV}}^{(i)}$.
    \STATE Update $\omega^{(\tau+1)}\leftarrow\omega^{(\tau)}-\alpha_\omega\nabla_\omega\mathcal L_Y(\omega^{(\tau)})$.
    \FOR{each $i\in\mathcal B$}
       \STATE Recompute $\log\widehat L_{\mathrm{IV}}^{(i)}$ as above with $\omega^{(\tau+1)}$ in place of $\omega^{(\tau)}$, using $M$ fresh draws $x^{(i,m)}\sim p_{\phi^{(\tau+1)}}(x\mid w^{(i)},\mathbf z_0^{(i,\tau)},\mathbf z_2^{(i,\tau)})$ when $X$ is continuous.
      \STATE Update the latent state by ascent on the objective in Eq.~\eqref{eq:latent_update_objective}:
      \STATE $\mathbf z^{(i,\tau+1)}\leftarrow\mathbf z^{(i,\tau)}+\beta_z\nabla_{\mathbf z^{(i,\tau)}}\big\{\log p(\mathbf z^{(i,\tau)})+\log p_{\theta^{(\tau+1)}}(\mathbf v^{(i)}\mid\mathbf z^{(i,\tau)})+\log p_{\phi^{(\tau+1)}}(x^{(i)}\mid w^{(i)},\mathbf z_0^{(i,\tau)},\mathbf z_2^{(i,\tau)})+\gamma\log\widehat L_{\mathrm{IV}}^{(i)}\big\}$.
    \ENDFOR
    \STATE Set $\mathbf z^{(i,\tau+1)}\leftarrow\mathbf z^{(i,\tau)}$ for $i\notin\mathcal B$, and $\tau\leftarrow\tau+1$.
  \ENDFOR
\ENDFOR
\STATE Return $(\theta,\phi,\omega)\leftarrow(\theta^{(\tau)},\phi^{(\tau)},\omega^{(\tau)})$.
\end{algorithmic}
\end{algorithm}

\begin{algorithm}[!ht]
\caption{Multi-start training and model selection}
\label{alg:causalbgm_iv_appendix}
\begin{algorithmic}[1]
\REQUIRE Training data $\mathcal D$; prior $p(\mathbf z)$; number of candidates $K$; initialization seeds $s_1,\ldots,s_K$; Monte Carlo size $M$; the inputs of Algorithms~\ref{alg:egm_appendix} and~\ref{alg:bgm_appendix}.
\ENSURE Selected generators $(\hat\theta,\hat\phi,\hat\omega)$.
\FOR{candidate $k=1,\ldots,K$}
  \STATE Run Algorithm~\ref{alg:egm_appendix} with seed $s_k$ to obtain $(\theta^{(0)},\phi^{(0)},\omega^{(0)})$ and $\{\mathbf z^{(i,0)}\}_{i=1}^n$.
  \STATE Run Algorithm~\ref{alg:bgm_appendix} from this initialization to obtain $(\theta_k,\phi_k,\omega_k)$.
  \FOR{$i=1,\ldots,n$}
    \STATE Compute $\hat{\mathbf z}^{(i)}=\arg\max_{\mathbf z}\{\log p(\mathbf z)+\log p_{\theta_k}(\mathbf v^{(i)}\mid\mathbf z)\}$ (Eq.~\eqref{eq:test_time_map}).
    \STATE Draw $x^{(i,m)}\sim p_{\phi_k}(x\mid w^{(i)},\hat{\mathbf z}_0^{(i)},\hat{\mathbf z}_2^{(i)})$, $m=1,\ldots,M$, and set $\bar\mu_k^{(i)}\leftarrow M^{-1}\sum_{m=1}^{M}\mu_{\omega_k}(x^{(i,m)},\hat{\mathbf z}_0^{(i)},\hat{\mathbf z}_1^{(i)})$.
  \ENDFOR
  \STATE Set $S_k\leftarrow n^{-1}\sum_{i=1}^{n}\{y^{(i)}-\bar\mu_k^{(i)}\}^2$ (Eq.~\eqref{eq:selection_criterion}).
\ENDFOR
\STATE Discard candidates with non-finite $S_k$; set $\hat k\leftarrow\arg\min_k S_k$, ties broken by the smallest index.
\STATE Return $(\hat\theta,\hat\phi,\hat\omega)\leftarrow(\theta_{\hat k},\phi_{\hat k},\omega_{\hat k})$.
\end{algorithmic}
\end{algorithm}

\begin{algorithm}[!ht]
\caption{Structural prediction and uncertainty quantification}
\label{alg:inference_appendix}
\begin{algorithmic}[1]
\REQUIRE Selected generators $(\hat\theta,\hat\phi,\hat\omega)$; evaluation pairs $\{(x_{\mathrm{eval}}^{(r)},\mathbf v_{\mathrm{eval}}^{(r)})\}_{r=1}^{R}$; benchmark structural values $g_0(x_{\mathrm{eval}}^{(r)},\mathbf v_{\mathrm{eval}}^{(r)})$; number of chains $C$, warm-up length $T_{\mathrm{warm}}$, and retained draws per chain $J_c$; interval level $1-\alpha$.
\ENSURE Structural predictions $\hat g(x_{\mathrm{eval}}^{(r)},\mathbf v_{\mathrm{eval}}^{(r)})$, structural MSE, and $100(1-\alpha)\%$ predictive intervals.
\STATE \textit{MAP readout.}
\FOR{test pair $(x_{\mathrm{eval}}^{(r)},\mathbf v_{\mathrm{eval}}^{(r)})$}
  \STATE Compute $\hat{\mathbf z}(\mathbf v_{\mathrm{eval}}^{(r)})=\arg\max_{\mathbf z}\{\log p(\mathbf z)+\log p_{\hat\theta}(\mathbf v_{\mathrm{eval}}^{(r)}\mid\mathbf z)\}$.
  \STATE Set $\hat g(x_{\mathrm{eval}}^{(r)},\mathbf v_{\mathrm{eval}}^{(r)})\leftarrow\mu_{\hat\omega}(x_{\mathrm{eval}}^{(r)},\hat{\mathbf z}_0(\mathbf v_{\mathrm{eval}}^{(r)}),\hat{\mathbf z}_1(\mathbf v_{\mathrm{eval}}^{(r)}))$.
\ENDFOR
\STATE Set $\mathrm{MSE}_{\mathrm{struct}}\leftarrow R^{-1}\sum_{r=1}^{R}\{\hat g(x_{\mathrm{eval}}^{(r)},\mathbf v_{\mathrm{eval}}^{(r)})-g_0(x_{\mathrm{eval}}^{(r)},\mathbf v_{\mathrm{eval}}^{(r)})\}^2$.
\STATE \textit{Posterior sampling.}
\FOR{each distinct covariate value $\mathbf v$ among $\{\mathbf v_{\mathrm{eval}}^{(r)}\}_{r=1}^{R}$}
  \STATE Run a short pilot of Hamiltonian Monte Carlo on $\log p(\mathbf z)+\log p_{\hat\theta}(\mathbf v\mid\mathbf z)$ to estimate a diagonal mass matrix.
  \STATE Run $C$ chains from overdispersed starting points with this mass matrix, each with $T_{\mathrm{warm}}$ warm-up iterations, during which the step size is adapted by dual averaging, followed by $J_c$ retained draws; pool the $J=C\,J_c$ draws $\{\mathbf z^{(j)}(\mathbf v)\}_{j=1}^{J}$.
\ENDFOR
\STATE \textit{Intervals.}
\FOR{test pair $(x_{\mathrm{eval}}^{(r)},\mathbf v_{\mathrm{eval}}^{(r)})$}
  \STATE For $j=1,\ldots,J$, with $\mathbf z^{(j)}=\mathbf z^{(j)}(\mathbf v_{\mathrm{eval}}^{(r)})$: set $m_j\leftarrow\mu_{\hat\omega}(x_{\mathrm{eval}}^{(r)},\mathbf z_0^{(j)},\mathbf z_1^{(j)})$ and $s_j^2\leftarrow\sigma_{\hat\omega}^2(x_{\mathrm{eval}}^{(r)},\mathbf z_0^{(j)},\mathbf z_1^{(j)})$.
    \STATE \textit{Predictive interval:} let $\hat F(y)=J^{-1}\sum_{j=1}^{J}\Phi\big((y-m_j)/s_j\big)$ be the CDF of the mixture in Eq.~\eqref{eq:posterior_predictive}; set $[\ell_\alpha,u_\alpha]\leftarrow[\hat F^{-1}(\alpha/2),\hat F^{-1}(1-\alpha/2)]$ by bisection.
\ENDFOR
\STATE Return $\{\hat g(x_{\mathrm{eval}}^{(r)},\mathbf v_{\mathrm{eval}}^{(r)})\}_{r=1}^{R}$, $\mathrm{MSE}_{\mathrm{struct}}$, and the intervals.
\end{algorithmic}
\end{algorithm}

\paragraph{Reproducibility of the multi-start procedure.}
All $K$ candidates are trained on the same training sample with the same optimization schedule. Each candidate receives its own network-initialization seed and its own seed for the Monte Carlo draws of the selection criterion, whereas the mini-batch order of the EGM initialization and the random stream of the stochastic optimization training are shared across candidates, so that candidates differ only in their initialization. Candidates whose criterion is not finite are excluded before the minimum is taken, and ties are broken by the smallest candidate index. The $K$ candidates are part of the estimator's computational budget and are not treated as independent repetitions; all repetitions reported in Section~\ref{sec:experiments} use independent training samples.

\FloatBarrier
\section{BGM-IV Implementation Details and Hyperparameters}
\label{app:bgm_iv_hyperparameters}

All BGM-IV results use the settings below, fixed within each benchmark and shared across sample sizes, confounding levels, and repetitions. Table~\ref{tab:bgm_iv_shared_hyperparameters} lists the settings common to the three benchmarks and Table~\ref{tab:bgm_iv_benchmark_hyperparameters} the benchmark-specific ones.

\begin{table}[!ht]
\centering
\caption{BGM-IV optimization and inference settings shared by all benchmarks.}
\label{tab:bgm_iv_shared_hyperparameters}
\begin{tabular}{@{}p{3.4cm}p{10.0cm}@{}}
\toprule
Component & Setting \\
\midrule
Optimizer & Adam, $\beta_1=0.9$, $\beta_2=0.99$; no weight decay \\
EGM initialization & 50{,}000 iterations, learning rate $2\times10^{-4}$, 5 discriminator steps per generator step, gradient penalty 10, latent reconstruction on \\
EGM outcome term & 8-node Gauss--Hermite quadrature; first-stage scale is a moving average (0.99) of the squared first-stage residual, capped at 1; variance-head penalty 0.001 \\
Stochastic optimization training & 200 epochs, learning rate $10^{-4}$ for $\theta$, $\phi$, $\omega$ and for the latent states \\
Tempering exponent & $\gamma=0.01$ \\
IV Monte Carlo & $M=1{,}000$ treatment draws, in training and at evaluation \\
Multi-start selection & $K=10$ candidates, selection by $S_k$ in Eq.~\eqref{eq:selection_criterion} \\
Test-time MAP & 1{,}000 Adam steps, learning rate $10^{-4}$ \\
Networks & $e_\psi$ and $p_\theta$: five hidden layers of 64 units; $p_\phi$ and $p_\omega$: hidden layers $[64,32,8]$ with mean and variance heads; discriminator: $[64,32,8]$ with batch normalization and tanh; LeakyReLU (slope 0.2) elsewhere \\
HMC & Pilot: 4 chains, 400 warm-up and 240 retained steps, 5 leapfrog steps in warm-up and 3, 5, or 7 in the retained steps, used only to set a diagonal mass matrix. Production: 4 overdispersed chains, dual-averaging step size (target acceptance 0.9); initial step sizes, production leapfrog steps, and chain lengths in Table~\ref{tab:bgm_iv_benchmark_hyperparameters} \\
\bottomrule
\end{tabular}
\end{table}

\begin{table}[!ht]
\centering
\caption{Benchmark-specific BGM-IV settings.}
\label{tab:bgm_iv_benchmark_hyperparameters}
\setlength{\tabcolsep}{5pt}
\begin{tabular}{@{}lccc@{}}
\toprule
 & Demand & Vector proxy & MNIST \\
\midrule
Latent partition $[\mathbf z_0,\mathbf z_1,\mathbf z_2,\mathbf z_3]$ & $[3,2,1,2]$ & $[3,2,1,2]$ & $[2,1,1,2]$ \\
Batch size & 64 & 64 & 32 \\
Model covariate $\mathbf v$ & $(T,S)$ & $(T,\mathrm{PCA}_k(\mathbf r))$ & $(T,\text{image})$ \\
$\dim(\mathbf v)$ & 2 & $1+k$ & 785 \\
Covariate likelihood & Gaussian & Gaussian & Gaussian / Bernoulli \\
Scale of $T$ & learned & learned & fixed, 0.1 \\
Covariate networks & MLP & MLP & CNN \\
Scaling of $x$, $y$ & training mean/SD & training mean/SD & DFIV constants \\
HMC initial step size & 0.05 & 0.05 & 0.02 \\
HMC leapfrog steps, warm-up & 5 & 5 & 31 \\
HMC leapfrog steps, retained & 3, 5, or 7 & 3, 5, or 7 & 7, 15, or 31 \\
HMC warm-up / draws per chain & 2{,}000 / 5{,}000 & 2{,}000 / 5{,}000 & 1{,}000 / 1{,}000 \\
\bottomrule
\end{tabular}
\end{table}

\paragraph{Architecture.}
The demand and vector-proxy benchmarks use the fully connected model of Table~\ref{tab:bgm_iv_shared_hyperparameters}; in the vector-proxy benchmark the $k$ PCA scores enter it directly as covariates. In the image benchmark, the image encoder has two $3\times3$ convolution layers with 64 filters, max pooling, dropout 0.1, dense layers of widths 128 and 64, and a final MLP with hidden units $[128,64]$; the decoder maps the latent state to a $7\times7\times128$ tensor, applies two transposed-convolution upsampling blocks and a final convolutional block, and outputs pixel logits, with a Bernoulli likelihood on intensities in $[0,1]$ and a separate Gaussian head for $T$. The treatment and outcome networks are the same in all benchmarks.

\paragraph{Preprocessing.}
In the demand benchmark, $x$, $y$, $w$, and the covariates are standardized with training-set means and standard deviations (standard deviations below $10^{-6}$ are replaced by 1). The vector-proxy benchmark standardizes $x$, $y$, $w$, and $T$ in the same way; the PCA is fitted on the training proxies only, and the $k$ scores are rescaled by one common factor so that their average variance equals that of the raw proxy, without per-component whitening, and then share one standardization scale. In the image benchmark we use the fixed scaling $x\mapsto(x-17.779)/3.7$ and $y\mapsto(y+292.1)/158.0$ of the DFIV code base, keep $T$ on its raw scale, and divide pixel intensities by 255. Structural MSE is always reported on the original outcome scale.

\FloatBarrier
\section{Baselines and Evaluation Details}
\label{app:baseline_selection}

KIV, DeepIV, and DFIV use the implementations in the DFIV code base~\citep{xu2021dfiv}\footnote{\url{https://github.com/liyuan9988/DeepFeatureIV}} (PyTorch~\citep{paszke2019pytorch}), and DML-IV\footnote{\url{https://github.com/shaodaqian/DML-IV}} and FIL\footnote{\url{https://github.com/meta-inf/fil}} use the implementations released by their authors, each with the hyperparameters of the configuration its implementation provides for the demand design. Within a repetition all methods receive the same training sample: the baselines draw it from the demand-design generator of the DFIV code base with the repetition index as random seed, and the BGM-IV generator reproduces that random stream, so the training arrays are identical across methods.

For each repetition, BGM-IV, DeepIV, DFIV, DML-IV, and FIL are each trained $K=10$ times on this sample. For DeepIV, DFIV, DML-IV, and FIL, the ten runs differ only in the seed of the model-side randomness: network initialization, mini-batch order, dropout, Monte Carlo draws, and the data partitions that are part of the estimator (the cross-fitting folds of DML-IV and the stage split of FIL; the stage split of DFIV is the fixed split of its implementation). The ten BGM-IV candidates differ only in their initialization, as described in Appendix~\ref{app:detailed_algorithm}. After training, each run is scored by the method's own criterion defined below, computed on the training sample with all networks in evaluation mode. The run with the smallest criterion is retained; runs whose criterion is not finite are discarded first, and ties are broken by the smallest run index. In every experiment, all ten runs of every method in every cell had a finite criterion, so no run was discarded. The structural MSE of each run on the evaluation grid is recorded but never used in the selection. KIV has no random initialization: its two regularization parameters are chosen by its own stage-wise validation on the two halves of its training split over a grid of 50 values, and it is fit once per repetition, it was run on the demand design and on the image-covariate benchmark.

All selection criteria are training-sample versions of the IV moment restriction of Section~\ref{sec:problem_setup}. For a candidate structural function $g$, a first-stage law $\widehat F(x\mid w,\mathbf v)$, and a set $\mathcal I$ of training units, write
\begin{equation}
\widehat R(g,\widehat F;\mathcal I)
=
\frac{1}{|\mathcal I|}\sum_{i\in\mathcal I}
\Big\{y^{(i)}-\int g(x,\mathbf v^{(i)})\,d\widehat F(x\mid w^{(i)},\mathbf v^{(i)})\Big\}^{2},
\label{eq:iv_moment_residual}
\end{equation}
the squared residual of the moment equation $\mathbb E[Y\mid W,V]=\int g(x,V)\,dP(x\mid W,V)$. The second stage of every baseline minimizes a version of Eq.~\eqref{eq:iv_moment_residual} with its own estimate of the inner conditional expectation, and its selection criterion is that same quantity recomputed after training. BGM-IV is trained with the negative log working likelihood of Section~\ref{sec:training}, and its selection criterion is the IV-moment residual of Eq.~\eqref{eq:selection_criterion}, evaluated at the covariate-only latent states. The criteria differ in how the inner integral is evaluated, in the subsample $\mathcal I$, and in whether $y^{(i)}$ or a cross-fitted regression of it is the target. Each criterion is computed in its implementation's own outcome units ($\tilde y$ denotes a rescaled outcome) and is compared only across runs of the same method on the same training sample.

\emph{BGM-IV.} For candidate $k$ with parameters $(\theta_k,\phi_k,\omega_k)$, let $\hat{\mathbf z}^{(i)}=\hat{\mathbf z}_k(\mathbf v^{(i)})$ denote its covariate-only MAP latent state for unit $i$ (Eq.~\eqref{eq:test_time_map}). The criterion is the training-set IV-moment residual
\begin{equation}
S_k
=
\frac{1}{n}\sum_{i=1}^{n}
\Big\{ y^{(i)} - \int \mu_{\omega_k}\big(x,\hat{\mathbf z}_0^{(i)},\hat{\mathbf z}_1^{(i)}\big)\,
p_{\phi_k}\big(x\mid w^{(i)},\hat{\mathbf z}_0^{(i)},\hat{\mathbf z}_2^{(i)}\big)\,dx \Big\}^2 ,
\label{eq:selection_criterion}
\end{equation}
that is, Eq.~\eqref{eq:iv_moment_residual} with $g(x,\mathbf v)=\mu_{\omega_k}(x,\hat{\mathbf z}_0(\mathbf v),\hat{\mathbf z}_1(\mathbf v))$, $\widehat F=p_{\phi_k}(\cdot\mid w,\hat{\mathbf z}_0(\mathbf v),\hat{\mathbf z}_2(\mathbf v))$, and $\mathcal I$ the full training sample; the integral is computed by Monte Carlo with $M=1{,}000$ draws from $p_{\phi_k}$. The selected model is $\hat k=\arg\min_k S_k$, and all point estimates and predictive intervals are computed from this single model. The criterion measures how well a candidate satisfies the IV moment restriction on the data it was fitted to; it uses no validation split, held-out sample, evaluation grid, or structural truth, and the latent states entering it are inferred from $\mathbf v$ alone.

\emph{DeepIV.} The first stage is a mixture density network $\widehat F_{\mathrm{MDN}}(x\mid w,\mathbf v)$ and the second stage a response network $g$. The criterion is the sampled second-stage loss that DeepIV minimizes,
\begin{equation}
\widehat R_{\mathrm{DeepIV}}
=
\frac{1}{n}\sum_{i=1}^{n}\frac{1}{S}\sum_{s=1}^{S}
\big\{\tilde y^{(i)}-g(a^{(i,s)},\mathbf v^{(i)})\big\}^{2},
\qquad a^{(i,s)}\sim\widehat F_{\mathrm{MDN}}(\cdot\mid w^{(i)},\mathbf v^{(i)}),
\end{equation}
with $S=100$ draws per unit (training uses two). Its expectation over the draws equals Eq.~\eqref{eq:iv_moment_residual} plus the average first-stage variance $n^{-1}\sum_i\mathrm{Var}_{\widehat F_{\mathrm{MDN}}}\{g(X,\mathbf v^{(i)})\mid w^{(i)},\mathbf v^{(i)}\}$, so it is the IV-moment residual in the form DeepIV optimizes.

\emph{DFIV.} DFIV learns treatment features $\varphi(x)$, instrument features $\psi(w,\mathbf v)$, and covariate features $\chi(\mathbf v)$; it fits the stage-1 ridge regression of $\varphi(x)$ on $\psi(w,\mathbf v)$ on one half $\mathcal I_1$ of its training split and, on the other half $\mathcal I_2$, fits $g(x,\mathbf v)=\hat{\mathbf w}^{\top}\{\varphi(x)\otimes\chi(\mathbf v)\}$ by ridge regression of $y$ on the predicted features $\widehat\Phi^{(i)}=\widehat{\mathbb E}[\varphi(X)\mid w^{(i)},\mathbf v^{(i)}]\otimes\chi(\mathbf v^{(i)})$. The criterion is its stage-2 objective at the solution,
\begin{equation}
\widehat R_{\mathrm{DFIV}}
=
\frac{1}{|\mathcal I_2|}\sum_{i\in\mathcal I_2}
\big\{y^{(i)}-\hat{\mathbf w}^{\top}\widehat\Phi^{(i)}\big\}^{2}
+\lambda_2\|\hat{\mathbf w}\|^{2},
\end{equation}
where $\lambda_2$ is its stage-2 ridge parameter. The first term is Eq.~\eqref{eq:iv_moment_residual} on $\mathcal I_2$, because $\hat{\mathbf w}^{\top}\widehat\Phi^{(i)}$ is the linear-in-features estimate of $\int g(x,\mathbf v^{(i)})\,d\widehat F(x\mid w^{(i)},\mathbf v^{(i)})$.

\emph{DML-IV.} DML-IV splits the training sample into ten folds $\mathcal I_1,\ldots,\mathcal I_{10}$ and, for each fold $k$, fits on the other nine an outcome regression $\hat m_{-k}(w,\mathbf v)$ of $Y$ on $(W,V)$ and a mixture-of-Gaussians treatment density $\widehat F_{-k}(x\mid w,\mathbf v)$; the response network $g$ is then fit on all folds. The criterion is its cross-fitted second-stage loss,
\begin{equation}
\widehat R_{\mathrm{DML\text{-}IV}}
=
\frac{1}{n}\sum_{k=1}^{10}\sum_{i\in\mathcal I_k}
\Big\{\hat m_{-k}(w^{(i)},\mathbf v^{(i)})-\frac{1}{S}\sum_{s=1}^{S}g(a^{(i,s)},\mathbf v^{(i)})\Big\}^{2},
\end{equation}
with $a^{(i,s)}\sim\widehat F_{-k}(\cdot\mid w^{(i)},\mathbf v^{(i)})$ and $S=100$ (training uses one draw per unit). It is Eq.~\eqref{eq:iv_moment_residual} with $y^{(i)}$ replaced by the cross-fitted estimate of $\mathbb E[Y\mid W,V]$, the double-machine-learning form of the same moment equation.

\emph{FIL.} FIL splits the training sample into two halves, learns on one half a random-feature embedding $\varphi(x)$ of the treatment and a stage-1 regression estimate $\widehat{\mathbb E}[\varphi(X)\mid w,\mathbf v]$, and on the other half fits $g(x,\mathbf v)=\hat{\mathbf w}(\mathbf v)^{\top}\varphi(x)$, with $\hat{\mathbf w}(\cdot)$ a network of the covariates, by minimizing $\{\tilde y-\hat{\mathbf w}(\mathbf v)^{\top}\widehat{\mathbb E}[\varphi(X)\mid w,\mathbf v]\}^{2}$ with early stopping on the half $\mathcal I_{\mathrm{val}}$ it holds out from that fit. The criterion is the validation loss of the state FIL returns,
\begin{equation}
\widehat R_{\mathrm{FIL}}
=
\min_{\text{epochs}}\;
\frac{1}{|\mathcal I_{\mathrm{val}}|}\sum_{i\in\mathcal I_{\mathrm{val}}}
\big\{\tilde y^{(i)}-\hat{\mathbf w}(\mathbf v^{(i)})^{\top}\widehat{\mathbb E}[\varphi(X)\mid w^{(i)},\mathbf v^{(i)}]\big\}^{2},
\end{equation}
which is Eq.~\eqref{eq:iv_moment_residual} on $\mathcal I_{\mathrm{val}}$; the minimum over epochs is FIL's own early-stopping rule, so the criterion is the score by which FIL itself selects its model. Table~\ref{tab:baseline_criteria} summarizes the five criteria.

\begin{table}[!ht]
\centering
\caption{Training-side selection criteria used in the multi-start protocol. Each criterion is each baseline's own second-stage objective, recomputed on training data after training, and the criterion of BGM-IV is the IV-moment residual of Eq.~\eqref{eq:selection_criterion}; all are versions of the IV-moment residual in Eq.~\eqref{eq:iv_moment_residual}.}
\label{tab:baseline_criteria}
\setlength{\tabcolsep}{4pt}
\begin{tabular}{@{}lp{5.6cm}p{2.6cm}p{3.4cm}@{}}
\toprule
Method & Estimate of $\int g\,d\widehat F$ & Target & Sample $\mathcal I$ \\
\midrule
BGM-IV & Monte Carlo, $1{,}000$ draws from $p_{\phi}$ at the covariate-only MAP latents & $y$ & training sample \\
DeepIV & Monte Carlo, $100$ draws from the mixture density network (sampled loss) & $\tilde y$ & training sample \\
DFIV & Linear in the stage-1 predicted features, plus the ridge penalty & $y$ & second-stage half \\
DML-IV & Monte Carlo, $100$ draws from the cross-fitted mixture density & $\hat m_{-k}(w,\mathbf v)$ & training sample, ten folds \\
FIL & Stage-1 regression of the treatment embedding & $\tilde y$ & validation half \\
\bottomrule
\end{tabular}
\end{table}

The predictive intervals of BGM-IV are the quantiles of the posterior predictive distribution in Eq.~\eqref{eq:posterior_predictive}, computed for the selected model of each repetition at every point of the evaluation grid. Their coverage is the probability that a new outcome at an evaluation point, drawn around the true structural value with the unit-variance disturbance of the design, falls inside the interval, averaged over the grid, and their width is measured in outcome units. None of the baseline implementations returns interval estimates on these benchmarks. FIL admits quasi-Bayesian uncertainty quantification for its kernelized second stage, but its extension to exogenous covariates, the variant applicable to the demand designs used here, uses a neural second stage and returns point estimates only, and its authors note that uncertainty quantification would in principle require a Bayesian neural network second stage, which is not part of their algorithm~\citep[Appendix~G]{wang2022fil}. Interval estimates are therefore reported for BGM-IV alone.

\FloatBarrier
\section{Benchmark Details}
\label{app:experiment_details}

\subsection{Low-Dimensional Demand Design}
\label{app:demand_details}

The low-dimensional demand benchmark follows the synthetic airline demand design used in DeepIV, KIV, and DFIV~\citep{hartford2017deepiv,singh2019kiv,xu2021dfiv}. Let $S$ denote customer group, $T$ time of year, $C$ the cost-shifter instrument, and $U$ an unobserved demand shock. We draw
\[
S\sim \mathrm{Unif}\{1,\ldots,7\},\qquad
T\sim \mathrm{Unif}[0,10],\qquad
C,U,\eta\sim \mathcal N(0,1),
\]
and set
\[
\psi(T)=2\left\{\frac{(T-5)^4}{600}+\exp[-4(T-5)^2]+\frac{T}{10}-2\right\}.
\]
For a confounding level $\rho$, the structural disturbance and price are
\[
\epsilon=\rho U+\sqrt{1-\rho^2}\eta,\qquad
P=25+(C+3)\psi(T)+U.
\]
The structural demand function and observed outcome are
\[
f_0(P,T,S)=100+(10+P)S \psi(T)-2P,\qquad
Y=f_0(P,T,S)+\epsilon.
\]
Thus $P$ is endogenous through $U$, while $C$ shifts price and is mean independent of the structural disturbance conditional on the observed covariates. The observed covariates are $V=(T,S)$. Evaluation uses the standard grid with 20 prices in $[10,25]$, 20 time points in $[0,10]$, and all seven customer groups, giving 2800 structural evaluation points.

\subsection{High-Dimensional Vector-Proxy Demand Design}
\label{app:vector_details}

The vector-proxy benchmark keeps the treatment, instrument, outcome, structural function, and evaluation grid fixed. Instead of observing the customer group directly, we create type-specific prototypes $\boldsymbol{\mu}_1,\ldots,\boldsymbol{\mu}_7\in\mathbb R^{784}$ using a fixed feature seed, and a unit with customer group $S$ observes the proxy vector
\[
\mathbf{r}=\boldsymbol{\mu}_S+\sigma_{\mathrm{rep}}\boldsymbol{\eta},\qquad \boldsymbol{\eta}\sim\mathcal N(0,\mathbf{I}_{784}),
\]
with $\sigma_{\mathrm{rep}}=0.5$. The same prototypes are used for all methods, and the evaluation grid attaches proxies to the same 20-by-20-by-7 combinations of price, time, and customer group. In each repetition the proxy is reduced to its first $k$ principal components, fitted on the training proxies only, with a deterministic sign convention and one common rescaling of the scores (Appendix~\ref{app:bgm_iv_hyperparameters}); the identical transformation is applied to the training data and to the evaluation grid and is given to every method as the covariate $(T,\mathrm{PCA}_k(\mathbf r))$. Each method uses its demand-design network and hyperparameters with the input layer widened to $1+k$ covariates. We use $\rho=0.5$, $k\in\{1,2,4,6,8,16,32,64,128\}$, and $n\in\{5{,}000,10{,}000,20{,}000,50{,}000\}$, with 20 repetitions per configuration.

\subsection{Image-Covariate Demand Design}
\label{app:image_details}

The image-covariate benchmark keeps the same treatment, instrument, outcome, structural function, and evaluation grid as the low-dimensional demand design. The only change is the observation of customer group: instead of exposing $S$ as a categorical covariate, we attach a randomly sampled MNIST digit image with label $S$~\citep{lecun1998mnist}. The observed covariate is therefore $\mathbf{v}=[T,\mathrm{image}_{784}]$ with $v_{\mathrm{dim}}=785$. Training images are sampled from the MNIST training split and evaluation images from the test split, while the structural grid remains the same 20-by-20-by-7 demand grid.

\FloatBarrier
\section{Complete Experiment Results}
\label{app:full_results}

\subsection{Demand Design}
\label{app:full_results_demand}

Table~\ref{tab:demand_mse} reports the structural MSE summarized in Figure~\ref{fig:demand_lowdim_box} for all six methods, three training sample sizes, and five confounding levels.

\begin{table}[!ht]
\centering
\caption{Structural MSE on the low-dimensional demand-design benchmark, grouped by training sample size $n$; entries are divided by $10^{2}$. Each cell reports the mean $\pm$ sample standard deviation over 20 repetitions, in which all methods use the same training data. Every method except KIV is selected among ten runs by its own training-side criterion (Appendix~\ref{app:baseline_selection}). The lowest mean in each row is in bold.}
\label{tab:demand_mse}
\setlength{\tabcolsep}{4pt}
\begin{tabular}{@{}lcccccc@{}}
\toprule
$\rho$ & DeepIV & KIV & DML-IV & FIL & DFIV & BGM-IV \\
\midrule
\multicolumn{7}{@{}l}{$n=1{,}000$} \\
\addlinespace[1pt]
0.1 & 51.52 $\pm$ 3.63 & 79.89 $\pm$ 20.02 & 36.23 $\pm$ 3.75 & 10.60 $\pm$ 5.18 & 10.22 $\pm$ 3.78 & \textbf{4.15 $\pm$ 0.91} \\
0.25 & 50.92 $\pm$ 3.70 & 73.74 $\pm$ 22.01 & 37.09 $\pm$ 3.94 & 11.11 $\pm$ 4.99 & 9.42 $\pm$ 1.87 & \textbf{3.83 $\pm$ 0.69} \\
0.5 & 51.04 $\pm$ 3.32 & 75.95 $\pm$ 18.77 & 35.88 $\pm$ 3.85 & 10.70 $\pm$ 4.82 & 10.05 $\pm$ 2.89 & \textbf{3.74 $\pm$ 0.47} \\
0.75 & 51.20 $\pm$ 3.60 & 73.62 $\pm$ 23.44 & 36.61 $\pm$ 4.26 & 10.36 $\pm$ 4.95 & 10.47 $\pm$ 4.41 & \textbf{4.36 $\pm$ 0.70} \\
0.9 & 51.14 $\pm$ 3.44 & 72.79 $\pm$ 21.36 & 36.06 $\pm$ 3.26 & 10.29 $\pm$ 4.90 & 10.25 $\pm$ 3.99 & \textbf{3.87 $\pm$ 0.69} \\
\midrule
\multicolumn{7}{@{}l}{$n=5{,}000$} \\
\addlinespace[1pt]
0.1 & 15.29 $\pm$ 1.89 & 51.16 $\pm$ 10.63 & 13.88 $\pm$ 1.73 & 2.50 $\pm$ 1.19 & 1.93 $\pm$ 0.32 & \textbf{1.72 $\pm$ 0.25} \\
0.25 & 15.53 $\pm$ 1.56 & 49.99 $\pm$ 13.53 & 13.92 $\pm$ 1.96 & 2.54 $\pm$ 1.10 & 1.95 $\pm$ 0.37 & \textbf{1.72 $\pm$ 0.37} \\
0.5 & 15.65 $\pm$ 1.96 & 47.49 $\pm$ 8.97 & 13.90 $\pm$ 2.09 & 2.37 $\pm$ 0.81 & 1.92 $\pm$ 0.29 & \textbf{1.81 $\pm$ 0.38} \\
0.75 & 15.59 $\pm$ 1.88 & 53.31 $\pm$ 11.02 & 14.62 $\pm$ 1.90 & 2.36 $\pm$ 0.81 & 1.84 $\pm$ 0.37 & \textbf{1.73 $\pm$ 0.39} \\
0.9 & 15.72 $\pm$ 2.00 & 49.87 $\pm$ 9.17 & 14.73 $\pm$ 1.85 & 2.32 $\pm$ 0.80 & 1.99 $\pm$ 0.31 & \textbf{1.70 $\pm$ 0.22} \\
\midrule
\multicolumn{7}{@{}l}{$n=10{,}000$} \\
\addlinespace[1pt]
0.1 & 11.42 $\pm$ 1.28 & 41.92 $\pm$ 8.58 & 11.27 $\pm$ 1.78 & 2.01 $\pm$ 0.49 & 1.50 $\pm$ 0.28 & \textbf{1.35 $\pm$ 0.18} \\
0.25 & 11.57 $\pm$ 1.53 & 39.51 $\pm$ 8.27 & 11.83 $\pm$ 1.70 & 2.01 $\pm$ 0.50 & 1.56 $\pm$ 0.26 & \textbf{1.38 $\pm$ 0.25} \\
0.5 & 11.65 $\pm$ 1.28 & 43.55 $\pm$ 11.29 & 11.20 $\pm$ 1.26 & 1.96 $\pm$ 0.46 & 1.51 $\pm$ 0.30 & \textbf{1.38 $\pm$ 0.26} \\
0.75 & 11.73 $\pm$ 1.53 & 41.04 $\pm$ 12.01 & 11.69 $\pm$ 2.26 & 1.93 $\pm$ 0.47 & 1.43 $\pm$ 0.34 & \textbf{1.38 $\pm$ 0.25} \\
0.9 & 11.80 $\pm$ 1.14 & 41.07 $\pm$ 7.23 & 11.62 $\pm$ 1.84 & 1.93 $\pm$ 0.46 & 1.44 $\pm$ 0.26 & \textbf{1.39 $\pm$ 0.31} \\
\bottomrule
\end{tabular}
\end{table}

\subsection{Vector-Proxy Benchmark}
\label{app:full_results_vector}

\begin{figure}[!ht]
  \centering
  \includegraphics[width=\linewidth]{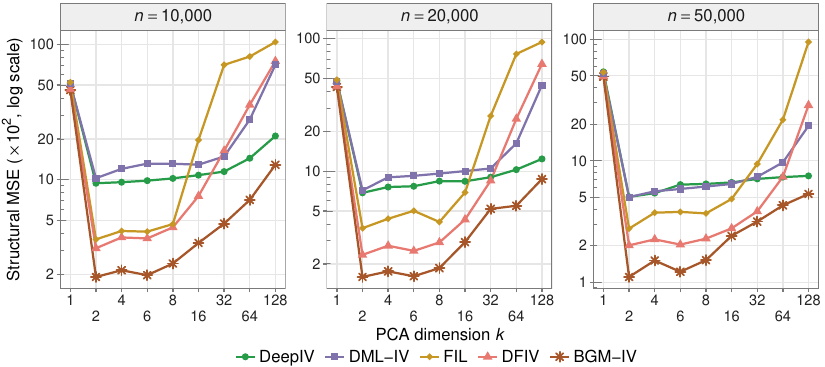}
  \caption{Structural MSE on the vector-proxy benchmark ($\rho=0.5$) at $n=10{,}000$, $20{,}000$, and $50{,}000$ as a function of the number $k$ of principal components, placed at equal spacing; each point is the mean over 20 repetitions, on a logarithmic axis in units of $10^{2}$. The corresponding panel for $n=5{,}000$ is Figure~\ref{fig:vector_mse_calibration}(a); numerical values, including standard deviations, are in Table~\ref{tab:vector_mse}.}
  \label{fig:vector_mse_by_n}
\end{figure}

Table~\ref{tab:vector_mse} reports the structural MSE of all methods for all four training sample sizes, and Figure~\ref{fig:vector_mse_by_n} shows the three sample sizes not displayed in Figure~\ref{fig:vector_mse_calibration}.

\begin{table}[!htp]
\centering
\caption{Structural MSE on the vector-proxy benchmark ($\rho=0.5$), grouped by training sample size $n$; rows give the number $k$ of principal components of the 784-dimensional proxy given to every method, and entries are divided by $10^{2}$. Each cell reports the mean $\pm$ sample standard deviation over 20 repetitions, in which all methods use the same training data; every method is selected among ten runs by its own training-side criterion (Appendix~\ref{app:baseline_selection}). The lowest mean in each row is in bold.}
\label{tab:vector_mse}
\setlength{\tabcolsep}{4pt}
\begin{tabular}{@{}lccccc@{}}
\toprule
$k$ & DeepIV & DML-IV & FIL & DFIV & BGM-IV \\
\midrule
\multicolumn{6}{@{}l}{$n=5{,}000$} \\
\addlinespace[1pt]
1 & 45.07 $\pm$ 28.57 & 46.95 $\pm$ 27.06 & 42.11 $\pm$ 29.31 & 38.60 $\pm$ 29.44 & \textbf{35.58 $\pm$ 27.87} \\
2 & 14.19 $\pm$ 7.75 & 15.82 $\pm$ 8.34 & 6.55 $\pm$ 7.85 & 6.61 $\pm$ 8.56 & \textbf{4.30 $\pm$ 6.86} \\
4 & 12.46 $\pm$ 1.10 & 15.32 $\pm$ 1.85 & 5.88 $\pm$ 1.56 & 5.44 $\pm$ 0.63 & \textbf{2.71 $\pm$ 0.35} \\
6 & 12.73 $\pm$ 1.40 & 16.32 $\pm$ 1.39 & 6.04 $\pm$ 2.04 & 5.81 $\pm$ 0.77 & \textbf{2.45 $\pm$ 0.50} \\
8 & 12.89 $\pm$ 1.20 & 16.04 $\pm$ 1.55 & 8.04 $\pm$ 2.85 & 7.16 $\pm$ 0.47 & \textbf{3.12 $\pm$ 0.53} \\
16 & 13.57 $\pm$ 1.32 & 17.29 $\pm$ 1.67 & 55.97 $\pm$ 23.99 & 13.24 $\pm$ 1.02 & \textbf{4.36 $\pm$ 0.89} \\
32 & 15.71 $\pm$ 2.08 & 22.61 $\pm$ 1.74 & 72.65 $\pm$ 5.47 & 24.83 $\pm$ 1.80 & \textbf{6.36 $\pm$ 0.71} \\
64 & 21.66 $\pm$ 1.14 & 54.84 $\pm$ 4.02 & 94.98 $\pm$ 5.22 & 44.76 $\pm$ 3.43 & \textbf{11.75 $\pm$ 2.55} \\
128 & \textbf{40.40 $\pm$ 2.83} & 100.07 $\pm$ 2.37 & 114.04 $\pm$ 4.49 & 77.53 $\pm$ 3.02 & 46.19 $\pm$ 8.27 \\
\midrule
\multicolumn{6}{@{}l}{$n=10{,}000$} \\
\addlinespace[1pt]
1 & 51.41 $\pm$ 30.37 & 51.12 $\pm$ 29.75 & 52.50 $\pm$ 31.33 & 46.52 $\pm$ 30.75 & \textbf{46.05 $\pm$ 30.36} \\
2 & 9.41 $\pm$ 1.16 & 10.25 $\pm$ 1.62 & 3.62 $\pm$ 0.92 & 3.11 $\pm$ 0.70 & \textbf{1.91 $\pm$ 0.36} \\
4 & 9.58 $\pm$ 1.15 & 12.02 $\pm$ 1.27 & 4.17 $\pm$ 0.89 & 3.74 $\pm$ 0.34 & \textbf{2.14 $\pm$ 0.22} \\
6 & 9.84 $\pm$ 0.94 & 13.15 $\pm$ 1.37 & 4.13 $\pm$ 1.23 & 3.68 $\pm$ 0.52 & \textbf{1.97 $\pm$ 0.20} \\
8 & 10.22 $\pm$ 1.04 & 13.12 $\pm$ 1.27 & 4.70 $\pm$ 1.34 & 4.44 $\pm$ 0.54 & \textbf{2.40 $\pm$ 0.31} \\
16 & 10.80 $\pm$ 0.98 & 12.90 $\pm$ 1.13 & 19.67 $\pm$ 15.61 & 7.56 $\pm$ 0.45 & \textbf{3.41 $\pm$ 0.43} \\
32 & 11.48 $\pm$ 1.10 & 14.85 $\pm$ 0.94 & 70.63 $\pm$ 4.52 & 16.36 $\pm$ 1.70 & \textbf{4.73 $\pm$ 0.56} \\
64 & 14.41 $\pm$ 1.87 & 27.81 $\pm$ 2.17 & 81.39 $\pm$ 5.75 & 35.77 $\pm$ 3.25 & \textbf{7.09 $\pm$ 0.83} \\
128 & 21.04 $\pm$ 1.91 & 70.67 $\pm$ 4.48 & 104.19 $\pm$ 3.31 & 75.26 $\pm$ 3.03 & \textbf{12.88 $\pm$ 1.38} \\
\midrule
\multicolumn{6}{@{}l}{$n=20{,}000$} \\
\addlinespace[1pt]
1 & 48.20 $\pm$ 33.94 & 47.52 $\pm$ 33.97 & 48.91 $\pm$ 35.41 & 43.80 $\pm$ 34.08 & \textbf{43.25 $\pm$ 33.99} \\
2 & 6.87 $\pm$ 1.00 & 7.17 $\pm$ 1.01 & 3.72 $\pm$ 1.18 & 2.34 $\pm$ 0.42 & \textbf{1.59 $\pm$ 0.44} \\
4 & 7.60 $\pm$ 0.97 & 8.99 $\pm$ 1.12 & 4.38 $\pm$ 1.26 & 2.74 $\pm$ 0.57 & \textbf{1.75 $\pm$ 0.25} \\
6 & 7.70 $\pm$ 0.89 & 9.23 $\pm$ 1.02 & 5.03 $\pm$ 1.49 & 2.50 $\pm$ 0.40 & \textbf{1.61 $\pm$ 0.31} \\
8 & 8.42 $\pm$ 0.80 & 9.60 $\pm$ 1.05 & 4.14 $\pm$ 1.79 & 2.90 $\pm$ 0.58 & \textbf{1.86 $\pm$ 0.24} \\
16 & 8.40 $\pm$ 0.87 & 10.03 $\pm$ 0.93 & 6.91 $\pm$ 1.88 & 4.32 $\pm$ 0.41 & \textbf{2.92 $\pm$ 0.40} \\
32 & 9.01 $\pm$ 1.11 & 10.49 $\pm$ 1.05 & 26.14 $\pm$ 6.10 & 8.50 $\pm$ 0.56 & \textbf{5.20 $\pm$ 2.56} \\
64 & 10.28 $\pm$ 1.25 & 16.24 $\pm$ 1.06 & 76.74 $\pm$ 7.00 & 24.84 $\pm$ 2.54 & \textbf{5.49 $\pm$ 0.65} \\
128 & 12.39 $\pm$ 1.26 & 44.70 $\pm$ 2.35 & 94.47 $\pm$ 6.59 & 64.51 $\pm$ 6.69 & \textbf{8.73 $\pm$ 1.13} \\
\midrule
\multicolumn{6}{@{}l}{$n=50{,}000$} \\
\addlinespace[1pt]
1 & 53.88 $\pm$ 25.16 & 52.06 $\pm$ 24.82 & 53.48 $\pm$ 24.89 & 49.74 $\pm$ 24.43 & \textbf{49.27 $\pm$ 25.15} \\
2 & 5.04 $\pm$ 0.47 & 5.03 $\pm$ 0.89 & 2.77 $\pm$ 0.99 & 2.01 $\pm$ 0.41 & \textbf{1.11 $\pm$ 0.17} \\
4 & 5.42 $\pm$ 0.62 & 5.60 $\pm$ 0.74 & 3.74 $\pm$ 1.14 & 2.26 $\pm$ 0.64 & \textbf{1.51 $\pm$ 0.22} \\
6 & 6.36 $\pm$ 0.67 & 5.86 $\pm$ 0.65 & 3.80 $\pm$ 1.33 & 2.04 $\pm$ 0.50 & \textbf{1.22 $\pm$ 0.18} \\
8 & 6.45 $\pm$ 0.66 & 6.13 $\pm$ 0.62 & 3.69 $\pm$ 1.20 & 2.29 $\pm$ 0.35 & \textbf{1.52 $\pm$ 0.26} \\
16 & 6.63 $\pm$ 0.87 & 6.44 $\pm$ 0.71 & 4.85 $\pm$ 1.31 & 2.78 $\pm$ 0.51 & \textbf{2.41 $\pm$ 0.58} \\
32 & 7.09 $\pm$ 0.85 & 7.37 $\pm$ 0.73 & 9.40 $\pm$ 4.75 & 3.83 $\pm$ 0.45 & \textbf{3.16 $\pm$ 0.32} \\
64 & 7.34 $\pm$ 0.78 & 9.69 $\pm$ 0.96 & 21.75 $\pm$ 4.89 & 7.37 $\pm$ 0.53 & \textbf{4.30 $\pm$ 1.13} \\
128 & 7.51 $\pm$ 0.72 & 19.36 $\pm$ 1.20 & 94.78 $\pm$ 6.58 & 28.62 $\pm$ 3.16 & \textbf{5.33 $\pm$ 1.00} \\
\bottomrule
\end{tabular}
\end{table}

\subsection{Image-Covariate Benchmark}
\label{app:full_results_mnist}

The structural MSE of all methods on the image-covariate benchmark is reported in Table~\ref{tab:mnist_main} of the main text. Figure~\ref{fig:mnist_curve} is drawn for one repetition of BGM-IV at $n=50{,}000$ and $\rho=0.5$, customer group $S=4$, and the grid price $P=17.11$, with time $T$ varying over the evaluation grid and each grid point carrying its own MNIST test image; the curve is the posterior mean of the structural function, the band is the 95\% predictive interval, and demand is normalized by the mean and standard deviation of the training outcomes. The intervals are quantiles of the posterior predictive distribution in Eq.~\eqref{eq:posterior_predictive} from 4 HMC chains with 1{,}000 warm-up and 1{,}000 retained draws each. Over the 20 repetitions of this setting, the 90\%, 95\%, and 99\% predictive intervals have mean coverage 90.7\%, 93.8\%, and 96.8\% and mean width 84.0, 99.5, and 129.7 outcome units, with standard deviations below 1 point for coverage and below 8 outcome units for width.

\subsection{Weak Instruments on the Image-Covariate Benchmark}
\label{app:full_results_weakiv}

Table~\ref{tab:mnist_weakiv} varies the strength of the instrument on the image-covariate benchmark.

\begin{table}[!ht]
\centering
\caption{Structural MSE on the image-covariate demand-design benchmark as the instrument strength $\lambda$ varies, grouped by training sample size $n$; entries are divided by $10^{2}$. The price equation is $P=25+(\lambda C+3)\psi(T)+U$, so $\lambda=1$ is the standard benchmark and smaller $\lambda$ weakens the instrument; the confounding level is $\rho=0.9$. Following \citet{shao2024dmliv}, the MSE is computed on 10{,}000 test points drawn from the same data-generating process, so that the test points follow the observed distribution of price and covariates rather than the fixed price grid. Each cell reports the mean $\pm$ sample standard deviation over 10 repetitions, in which all methods use the same training data; every method is selected among ten runs by its own training-side criterion (Appendix~\ref{app:baseline_selection}). The lowest mean in each row is in bold.}
\label{tab:mnist_weakiv}
\setlength{\tabcolsep}{4pt}
\begin{tabular}{@{}lccccc@{}}
\toprule
$\lambda$ & DeepIV & DML-IV & FIL & DFIV & BGM-IV \\
\midrule
\multicolumn{6}{@{}l}{$n=5{,}000$} \\
\addlinespace[1pt]
1 & 52.73 $\pm$ 3.22 & 53.39 $\pm$ 1.71 & 70.38 $\pm$ 6.91 & 69.30 $\pm$ 3.32 & \textbf{19.85 $\pm$ 1.69} \\
0.8 & 45.09 $\pm$ 2.31 & 49.80 $\pm$ 2.10 & 63.51 $\pm$ 8.26 & 71.01 $\pm$ 2.76 & \textbf{20.40 $\pm$ 1.40} \\
0.6 & 39.69 $\pm$ 2.21 & 44.77 $\pm$ 2.28 & 56.59 $\pm$ 2.80 & 68.95 $\pm$ 3.58 & \textbf{20.91 $\pm$ 1.62} \\
0.4 & 33.99 $\pm$ 1.60 & 40.71 $\pm$ 2.21 & 49.37 $\pm$ 4.94 & 68.30 $\pm$ 4.38 & \textbf{22.57 $\pm$ 2.44} \\
0.2 & 29.02 $\pm$ 1.34 & 37.72 $\pm$ 2.52 & 43.15 $\pm$ 4.14 & 67.54 $\pm$ 4.00 & \textbf{26.74 $\pm$ 3.68} \\
\midrule
\multicolumn{6}{@{}l}{$n=10{,}000$} \\
\addlinespace[1pt]
1 & 41.27 $\pm$ 2.04 & 44.70 $\pm$ 2.07 & 60.52 $\pm$ 2.92 & 55.76 $\pm$ 3.78 & \textbf{16.26 $\pm$ 1.50} \\
0.8 & 35.44 $\pm$ 1.87 & 42.08 $\pm$ 1.81 & 57.97 $\pm$ 2.89 & 54.48 $\pm$ 4.69 & \textbf{15.14 $\pm$ 1.31} \\
0.6 & 29.71 $\pm$ 1.96 & 38.60 $\pm$ 1.31 & 49.66 $\pm$ 2.81 & 52.96 $\pm$ 4.76 & \textbf{16.29 $\pm$ 1.32} \\
0.4 & 24.73 $\pm$ 1.20 & 34.37 $\pm$ 1.59 & 42.91 $\pm$ 4.13 & 53.06 $\pm$ 6.12 & \textbf{18.33 $\pm$ 2.51} \\
0.2 & 21.34 $\pm$ 1.61 & 31.81 $\pm$ 1.43 & 35.85 $\pm$ 3.40 & 52.12 $\pm$ 4.49 & \textbf{18.95 $\pm$ 1.90} \\
\bottomrule
\end{tabular}
\end{table}

\FloatBarrier
\section{Paired Comparisons Against the Strongest Baseline}
\label{app:pvalue_comparisons}

\begin{table}[!htp]
\centering
\caption{Paired comparison of BGM-IV with the strongest baseline on the vector-proxy benchmark ($\rho=0.5$), grouped by training sample size $n$; $k$ is the number of principal components. In every configuration the strongest baseline is the one with the lowest mean structural MSE (Table~\ref{tab:vector_mse}). ``Baseline'' and ``BGM-IV'' are mean structural MSEs over 20 repetitions divided by $10^{2}$; ``Difference'' is the mean paired difference, BGM-IV minus baseline, in the same units, so negative values favor BGM-IV; ``Lower in'' is the number of repetitions in which BGM-IV has the lower error. The last column is the two-sided exact Wilcoxon signed-rank $p$-value, Holm-adjusted over the 36 configurations of the table.}
\label{tab:paired_vector}
\setlength{\tabcolsep}{5pt}
\begin{tabular}{@{}llccccc@{}}
\toprule
$k$ & Strongest baseline & Baseline & BGM-IV & Difference & Lower in & $p$ (Holm) \\
\midrule
\multicolumn{7}{@{}l}{$n=5{,}000$} \\
\addlinespace[1pt]
1 & DFIV & 38.60 & 35.58 & $-3.02$ & 17/20 & 0.005 \\
2 & FIL & 6.55 & 4.30 & $-2.25$ & 19/20 & $<0.001$ \\
4 & DFIV & 5.44 & 2.71 & $-2.73$ & 20/20 & $<0.001$ \\
6 & DFIV & 5.81 & 2.45 & $-3.36$ & 20/20 & $<0.001$ \\
8 & DFIV & 7.16 & 3.12 & $-4.04$ & 20/20 & $<0.001$ \\
16 & DFIV & 13.24 & 4.36 & $-8.88$ & 20/20 & $<0.001$ \\
32 & DeepIV & 15.71 & 6.36 & $-9.35$ & 20/20 & $<0.001$ \\
64 & DeepIV & 21.66 & 11.75 & $-9.91$ & 20/20 & $<0.001$ \\
128 & DeepIV & 40.40 & 46.19 & $+5.79$ & 3/20 & 0.036 \\
\midrule
\multicolumn{7}{@{}l}{$n=10{,}000$} \\
\addlinespace[1pt]
1 & DFIV & 46.52 & 46.05 & $-0.47$ & 13/20 & 0.231 \\
2 & DFIV & 3.11 & 1.91 & $-1.20$ & 19/20 & $<0.001$ \\
4 & DFIV & 3.74 & 2.14 & $-1.60$ & 20/20 & $<0.001$ \\
6 & DFIV & 3.68 & 1.97 & $-1.71$ & 20/20 & $<0.001$ \\
8 & DFIV & 4.44 & 2.40 & $-2.04$ & 20/20 & $<0.001$ \\
16 & DFIV & 7.56 & 3.41 & $-4.16$ & 20/20 & $<0.001$ \\
32 & DeepIV & 11.48 & 4.73 & $-6.74$ & 20/20 & $<0.001$ \\
64 & DeepIV & 14.41 & 7.09 & $-7.32$ & 20/20 & $<0.001$ \\
128 & DeepIV & 21.04 & 12.88 & $-8.16$ & 20/20 & $<0.001$ \\
\midrule
\multicolumn{7}{@{}l}{$n=20{,}000$} \\
\addlinespace[1pt]
1 & DFIV & 43.80 & 43.25 & $-0.55$ & 15/20 & 0.160 \\
2 & DFIV & 2.34 & 1.59 & $-0.75$ & 18/20 & 0.001 \\
4 & DFIV & 2.74 & 1.75 & $-0.98$ & 20/20 & $<0.001$ \\
6 & DFIV & 2.50 & 1.61 & $-0.89$ & 18/20 & $<0.001$ \\
8 & DFIV & 2.90 & 1.86 & $-1.04$ & 19/20 & $<0.001$ \\
16 & DFIV & 4.32 & 2.92 & $-1.40$ & 20/20 & $<0.001$ \\
32 & DFIV & 8.50 & 5.20 & $-3.30$ & 17/20 & 0.001 \\
64 & DeepIV & 10.28 & 5.49 & $-4.79$ & 20/20 & $<0.001$ \\
128 & DeepIV & 12.39 & 8.73 & $-3.66$ & 19/20 & $<0.001$ \\
\midrule
\multicolumn{7}{@{}l}{$n=50{,}000$} \\
\addlinespace[1pt]
1 & DFIV & 49.74 & 49.27 & $-0.47$ & 16/20 & 0.061 \\
2 & DFIV & 2.01 & 1.11 & $-0.91$ & 20/20 & $<0.001$ \\
4 & DFIV & 2.26 & 1.51 & $-0.74$ & 19/20 & $<0.001$ \\
6 & DFIV & 2.04 & 1.22 & $-0.82$ & 20/20 & $<0.001$ \\
8 & DFIV & 2.29 & 1.52 & $-0.77$ & 20/20 & $<0.001$ \\
16 & DFIV & 2.78 & 2.41 & $-0.37$ & 15/20 & 0.160 \\
32 & DFIV & 3.83 & 3.16 & $-0.67$ & 20/20 & $<0.001$ \\
64 & DeepIV & 7.34 & 4.30 & $-3.04$ & 18/20 & $<0.001$ \\
128 & DeepIV & 7.51 & 5.33 & $-2.18$ & 19/20 & $<0.001$ \\
\bottomrule
\end{tabular}
\end{table}

For the two high-dimensional benchmarks, we compare BGM-IV in every configuration with the baseline that has the lowest mean structural MSE in that configuration. Because all methods are fit to the same training sample within a repetition, the 20 repetitions form pairs, and we test the paired differences with the two-sided exact Wilcoxon signed-rank test, adjusting the $p$-values by Holm's procedure~\citep{holm1979simple} within each benchmark. Tables~\ref{tab:paired_vector} and~\ref{tab:paired_mnist} report the results. On the vector-proxy benchmark, BGM-IV is significantly more accurate than the strongest baseline at the 5\% level in 31 of the 36 configurations. Four configurations show no significant difference: $k=1$ for $n\ge10{,}000$, where no method recovers the customer type, and $(n=50{,}000,\,k=16)$. In the remaining configuration, $(n=5{,}000,\,k=128)$, DeepIV is significantly more accurate. On the image-covariate benchmark, BGM-IV is significantly more accurate in all 15 configurations, and it has the lower error in every repetition for $n\ge5{,}000$.

\begin{table}[!htp]
\centering
\caption{Paired comparison of BGM-IV with the strongest baseline on the image-covariate benchmark, grouped by training sample size $n$. In every configuration the strongest baseline is the one with the lowest mean structural MSE (Table~\ref{tab:mnist_main}). ``Baseline'' and ``BGM-IV'' are mean structural MSEs over 20 repetitions divided by $10^{2}$; ``Difference'' is the mean paired difference, BGM-IV minus baseline, in the same units, so negative values favor BGM-IV; ``Lower in'' is the number of repetitions in which BGM-IV has the lower error. The last column is the two-sided exact Wilcoxon signed-rank $p$-value, Holm-adjusted over the 15 configurations of the table.}
\label{tab:paired_mnist}
\setlength{\tabcolsep}{5pt}
\begin{tabular}{@{}llccccc@{}}
\toprule
$\rho$ & Strongest baseline & Baseline & BGM-IV & Difference & Lower in & $p$ (Holm) \\
\midrule
\multicolumn{7}{@{}l}{$n=1{,}000$} \\
\addlinespace[1pt]
0.1 & DFIV & 102.65 & 80.04 & $-22.62$ & 17/20 & $<0.001$ \\
0.25 & DFIV & 102.92 & 79.18 & $-23.74$ & 18/20 & $<0.001$ \\
0.5 & DFIV & 103.20 & 75.85 & $-27.36$ & 19/20 & $<0.001$ \\
0.75 & DFIV & 102.90 & 80.50 & $-22.40$ & 20/20 & $<0.001$ \\
0.9 & DFIV & 101.58 & 78.36 & $-23.22$ & 20/20 & $<0.001$ \\
\midrule
\multicolumn{7}{@{}l}{$n=5{,}000$} \\
\addlinespace[1pt]
0.1 & DFIV & 69.11 & 26.50 & $-42.61$ & 20/20 & $<0.001$ \\
0.25 & DFIV & 71.22 & 25.32 & $-45.90$ & 20/20 & $<0.001$ \\
0.5 & DFIV & 71.10 & 25.54 & $-45.56$ & 20/20 & $<0.001$ \\
0.75 & DFIV & 71.12 & 24.91 & $-46.21$ & 20/20 & $<0.001$ \\
0.9 & DFIV & 69.33 & 25.94 & $-43.39$ & 20/20 & $<0.001$ \\
\midrule
\multicolumn{7}{@{}l}{$n=10{,}000$} \\
\addlinespace[1pt]
0.1 & DFIV & 55.38 & 17.12 & $-38.25$ & 20/20 & $<0.001$ \\
0.25 & DFIV & 55.82 & 17.16 & $-38.66$ & 20/20 & $<0.001$ \\
0.5 & DFIV & 55.73 & 17.14 & $-38.58$ & 20/20 & $<0.001$ \\
0.75 & DFIV & 55.68 & 17.13 & $-38.54$ & 20/20 & $<0.001$ \\
0.9 & DFIV & 54.69 & 17.14 & $-37.54$ & 20/20 & $<0.001$ \\
\bottomrule
\end{tabular}
\end{table}

\FloatBarrier
\section{Computational Cost}
\label{app:computational_cost}

\begin{figure}[!ht]
  \centering
  \includegraphics[width=\linewidth]{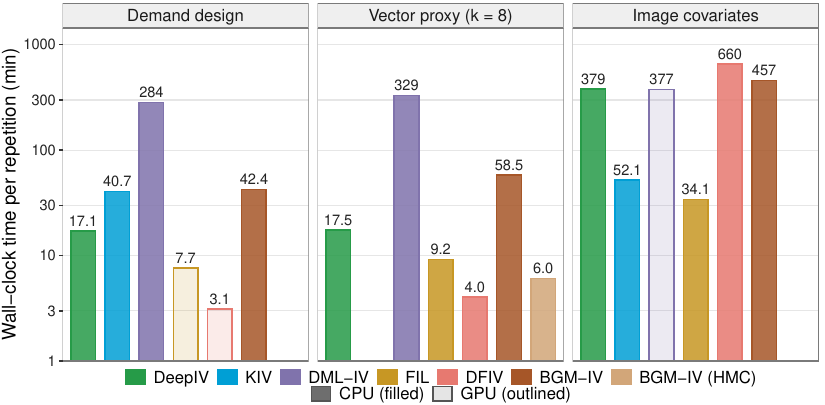}
  \caption{Median wall-clock time of one repetition at $n=10{,}000$ on the three benchmarks, on a logarithmic axis in minutes, with the value printed above each bar. Filled bars were run on CPU and outlined bars on GPU. BGM-IV (HMC) is the separate sampling job for the predictive intervals, shown for the vector-proxy benchmark; the corresponding job of the image-covariate benchmark was run at $n=50{,}000$ (Table~\ref{tab:compute_cost}).}
  \label{fig:compute_cost}
\end{figure}

All BGM-IV models were trained on CPU nodes of a shared computing cluster; no GPU was used for training on any benchmark. The HMC sampler of Section~\ref{sec:model_inference} ran on CPU for the vector-proxy benchmark, where one sampling job took between 5 and 9 minutes per repetition, and on a GPU for the image-covariate benchmark. The baselines ran on the devices supported by their implementations, which differ across methods and sample sizes. Table~\ref{tab:compute_cost} and Figure~\ref{fig:compute_cost} report the median wall-clock time of one repetition, that is, the ten training runs of the multi-start protocol together with the selection step and the evaluation, and, for BGM-IV, the separate wall-clock time of one HMC job. We report $n=10{,}000$ as a representative setting. On the image-covariate benchmark the predictive intervals are computed at $n=50{,}000$ (Figure~\ref{fig:mnist_curve}), where training took 1{,}244 minutes on CPU and one HMC job 67 minutes on a GPU per repetition. On the vector-proxy benchmark the times at $k=128$ are close to those at $k=8$ (49 instead of 59 minutes for BGM-IV). Because the devices differ, these timings document the cost of the reported experiments and are not a controlled speed comparison.

\begin{table}[!ht]
\centering
\caption{Median wall-clock time in minutes of one repetition (ten training runs of the multi-start protocol, selection, and evaluation) and, for BGM-IV, of one HMC job (4 chains; 2{,}000 warm-up and 5{,}000 retained draws per chain), with the device in parentheses. Medians are taken over the repetitions of all confounding levels at the given sample size. HMC is reported for the vector-proxy benchmark; on the image-covariate benchmark it was run at $n=50{,}000$ (see the text), and it is not reported for the demand design. KIV has no multi-start protocol, so its entry is the time of a single fit; KIV was not run on the vector-proxy benchmark.}
\label{tab:compute_cost}
\setlength{\tabcolsep}{6pt}
\begin{tabular}{@{}lccc@{}}
\toprule
 & Demand & Vector proxy ($k=8$) & Image \\
Method & $n=10{,}000$ & $n=10{,}000$ & $n=10{,}000$ \\
\midrule
KIV & 40.7 (CPU) & -- & 52.1 (CPU) \\
DeepIV & 17.1 (CPU) & 17.5 (CPU) & 379 (CPU) \\
DML-IV & 284 (CPU) & 329 (CPU) & 377 (GPU) \\
FIL & 7.7 (GPU) & 9.2 (CPU) & 34.1 (CPU) \\
DFIV & 3.1 (GPU) & 4.0 (CPU) & 660 (CPU) \\
\midrule
BGM-IV, training & 42.4 (CPU) & 58.5 (CPU) & 457 (CPU) \\
BGM-IV, HMC & -- & 6.0 (CPU) & -- \\
\bottomrule
\end{tabular}
\end{table}

\FloatBarrier
\section{Ablation Details}
\label{app:ablation}

This appendix describes the settings and variants of the ablation study of Section~\ref{sec:ablation}, reports the paired comparisons, and compares BGM-IV with the two strongest baselines in every setting. All runs use the low-dimensional demand design of Appendix~\ref{app:demand_details} at $n=1{,}000$ and $\rho\in\{0.1,0.5,0.9\}$, with 20 repetitions per configuration. Each variant changes one component of BGM-IV and keeps everything else unchanged, namely the hyperparameters of Appendix~\ref{app:bgm_iv_hyperparameters} including the tempering exponent $\gamma=0.01$, the $K=10$ candidates and the selection rule of Eq.~\eqref{eq:selection_criterion}, the covariate-only MAP readout, and the training sample and initialization seeds of each repetition. Every variant is therefore paired with the full model by repetition. The structural MSE is evaluated on the 2800-point grid of Appendix~\ref{app:demand_details}. Table~\ref{tab:ablation_main} of the main text reports the mean and standard deviation over repetitions, and Table~\ref{tab:ablation_paired} the ratio of the mean structural MSE of each variant to that of the full model, together with the number of repetitions in which the full model has the lower error.

\paragraph{Settings.}
Each component is examined in a setting in which the difficulty it addresses is present. \emph{Standard}: the design of Appendix~\ref{app:demand_details} without modification. \emph{Amplified confounding}: the structural disturbance is amplified by a factor of 100, $\epsilon=100\,(\rho U+\sqrt{1-\rho^2}\,\eta)$. The price equation, the instrument, the structural function $f_0$, and the evaluation grid are unchanged, and the correlation between $\epsilon$ and the confounder $U$ remains $\rho$. \emph{Nuisance covariates}: five variables $N_1,\ldots,N_5\sim\mathcal N(0,1)$, independent of each other and of all other variables, are appended to the observed covariates, $V=(T,S,N_1,\ldots,N_5)$. They enter neither the price nor the outcome, so the structural function is unchanged. In each repetition they are drawn after all other variables from the same random stream, so that the remaining variables coincide with those of the standard design, and the evaluation grid receives a single fixed draw. The covariate dimension is $\dim(\mathbf v)=7$. In the standard and amplified-confounding settings the latent partition is $[\mathbf z_0,\mathbf z_1,\mathbf z_2,\mathbf z_3]=[3,2,1,2]$ as in Appendix~\ref{app:bgm_iv_hyperparameters}; the partition of the nuisance-covariate setting is given below.

\paragraph{Choice of the amplification.}
The amplification is chosen so that the confounding bias is clearly visible at $n=1{,}000$. In the demand design, $P-25-3\psi(T)=C\psi(T)+U$ with $C$ and $U$ independent standard normal, so $\mathbb E[U\mid P,T,S]=(P-25-3\psi(T))/(1+\psi(T)^{2})$, and a regression of the outcome on the observed price and the covariates converges to $f_0+\mathbb E[\epsilon\mid P,T,S]$ with $\mathbb E[\epsilon\mid P,T,S]=c\rho\,\mathbb E[U\mid P,T,S]$, where $c$ is the amplification factor. The squared confounding bias of such a regression therefore scales with $c^{2}\rho^{2}$. Amplifying the disturbance by $c=100$ makes this bias large relative to the estimation error at $n=1{,}000$, so that the two likelihoods can be clearly distinguished, and the bias grows with $\rho^{2}$, in the order of the increases in Table~\ref{tab:ablation_main}.

\paragraph{Without EGM initialization (standard setting).}
The variant reduces the EGM initialization of Algorithm~\ref{alg:egm_appendix} to a single minibatch update, so the generators and the encoder enter the stochastic optimization training essentially at their random initialization, and the latent states are set from this untrained encoder. The stochastic optimization training, the selection, and the readout are unchanged. Removing the initialization increases the mean structural MSE by a factor of 17.7 to 18.9, and the full model has the lower error in all 60 paired repetitions.

\paragraph{Observed-treatment likelihood (amplified-confounding setting).}
The variant replaces the IV-integrated pseudo-likelihood $p_{\mathrm{IV}}(y\mid w,\mathbf z_0,\mathbf z_1,\mathbf z_2)$ by the outcome likelihood $p_\omega(y\mid x,\mathbf z_0,\mathbf z_1)$ evaluated at the observed treatment, in the update of the outcome generator in Eq.~\eqref{eq:loss_y_batch} and in the latent update in Eq.~\eqref{eq:latent_update_objective}, with the same tempering exponent $\gamma=0.01$. Correspondingly, the integrated outcome moment of the EGM initialization in Eq.~\eqref{eq:egm_outcome_moment} is replaced by the outcome mean at the observed treatment. The instrument still enters the treatment model, which is trained as in the full model, and the latent partition, the selection, and the readout are retained. The structural MSE increases by a factor of 1.18, 1.98, and 3.63 at $\rho=0.1$, $0.5$, and $0.9$, and the full model has the lower error in 15, 20, and 20 of the 20 paired repetitions.

\paragraph{Unstructured latent space (nuisance-covariate setting).}
In this setting the block sizes of the full model are specified from the design of the covariates: one dimension for $\mathbf z_0$, since the time of year enters both price and demand, one for $\mathbf z_1$, since the customer group enters demand only, none for $\mathbf z_2$, since no covariate enters price only, and five for the nuisance block $\mathbf z_3$, one per nuisance covariate, giving $[\mathbf z_0,\mathbf z_1,\mathbf z_2,\mathbf z_3]=[1,1,0,5]$. Only the block sizes are specified; the assignment of covariates to blocks is learned. The unstructured variant assigns all seven latent dimensions to a single block that enters every generator, so that the covariate generator takes $\mathbf z$, the treatment generator takes $(w,\mathbf z)$, and the outcome generator takes $(x,\mathbf z)$. The total latent dimension, the IV-integrated pseudo-likelihood, the EGM initialization, and all other components are retained. The unstructured latent vector increases the structural MSE by a factor of 1.40, 1.44, and 1.43 at $\rho=0.1$, $0.5$, and $0.9$, and the full model has the lower error in 17, 18, and 17 of the 20 paired repetitions.

\paragraph{Location of the nuisance information.}
To locate the information about the nuisance covariates in the trained latent space, we regress each nuisance covariate on the latent states of the training observations at the end of training, for the selected candidate of each repetition, and report the five-fold cross-validated $R^{2}$ of a linear regression, averaged over the five covariates and the 60 repetitions. In the partitioned model, the nuisance block $\mathbf z_3$ attains $R^{2}=0.96$, whereas the coordinates read by the treatment and outcome generators, $\mathbf z_0$ and $(\mathbf z_0,\mathbf z_1)$, attain $R^{2}\approx 0$ ($-0.01$). In the unstructured model, every generator reads all coordinates, which attain $R^{2}=0.96$. The partition therefore keeps the nuisance variation out of the treatment and outcome generators.

\paragraph{Comparison with baselines.}
To put the error levels of the modified settings in context, we also run DFIV and FIL, the two strongest baselines in the standard design at $n=1{,}000$ (Table~\ref{tab:demand_mse}), in all three settings with the protocol of Appendix~\ref{app:baseline_selection}. In the nuisance-covariate setting, the five covariates are appended to the covariates of each baseline, and to its instrument vector where that vector contains the covariates, and the input layers are widened accordingly. Table~\ref{tab:ablation_baselines} reports the results. Both modifications make the problem harder for every method; for example, relative to the standard design, the structural MSE of DFIV increases by a factor of 7.4 to 9.9 under amplified confounding and 7.9 to 8.3 with the nuisance covariates, and that of BGM-IV by a factor of 8.2 to 9.0 and 7.8 to 8.8. BGM-IV attains the lowest mean structural MSE in every setting and at every confounding level, and has the lower error than each baseline in at least 19 of the 20 paired repetitions of every configuration.

\begin{table}[!ht]
\centering
\caption{Paired comparison of each ablation variant with the full model on the demand design at $n=1{,}000$. Each cell reports the ratio of the mean structural MSE of the variant to that of the full model, and in parentheses the number of repetitions, out of 20, in which the full model has the lower error.}
\label{tab:ablation_paired}
\footnotesize
\setlength{\tabcolsep}{6pt}
\begin{tabular}{@{}ll ccc@{}}
\toprule
Setting & Variant & $\rho=0.1$ & $\rho=0.5$ & $\rho=0.9$ \\
\midrule
Standard & without EGM initialization & 17.67 (20) & 18.85 (20) & 17.88 (20) \\
Amplified confounding & observed-treatment likelihood & 1.18 (15) & 1.98 (20) & 3.63 (20) \\
Nuisance covariates & unstructured latent space & 1.40 (17) & 1.44 (18) & 1.43 (17) \\
\bottomrule
\end{tabular}
\end{table}

\begin{table}[!ht]
\centering
\caption{Structural MSE (divided by $10^{2}$, mean $\pm$ standard deviation over 20 repetitions) of BGM-IV and of DFIV and FIL, the two strongest baselines in the standard design, in the three ablation settings on the demand design at $n=1{,}000$. BGM-IV is the full model of Table~\ref{tab:ablation_main} in each setting, and the lowest mean in each setting and column is in bold.}
\label{tab:ablation_baselines}
\footnotesize
\setlength{\tabcolsep}{6pt}
\begin{tabular}{@{}ll ccc@{}}
\toprule
Setting & Method & $\rho=0.1$ & $\rho=0.5$ & $\rho=0.9$ \\
\midrule
\multirow{3}{*}{Standard} & BGM-IV & \textbf{4.15 $\pm$ 0.91} & \textbf{3.74 $\pm$ 0.47} & \textbf{3.87 $\pm$ 0.69} \\
 & DFIV & 10.22 $\pm$ 3.78 & 10.05 $\pm$ 2.89 & 10.25 $\pm$ 3.99 \\
 & FIL & 10.60 $\pm$ 5.18 & 10.70 $\pm$ 4.82 & 10.29 $\pm$ 4.90 \\
\midrule
\multirow{3}{*}{Amplified confounding} & BGM-IV & \textbf{35.21 $\pm$ 8.03} & \textbf{33.82 $\pm$ 9.01} & \textbf{31.65 $\pm$ 6.23} \\
 & DFIV & 101.23 $\pm$ 44.32 & 94.64 $\pm$ 38.53 & 76.02 $\pm$ 17.60 \\
 & FIL & 59.40 $\pm$ 23.75 & 58.50 $\pm$ 23.07 & 64.26 $\pm$ 33.37 \\
\midrule
\multirow{3}{*}{Nuisance covariates} & BGM-IV & \textbf{32.34 $\pm$ 9.98} & \textbf{32.86 $\pm$ 9.89} & \textbf{30.53 $\pm$ 8.69} \\
 & DFIV & 82.15 $\pm$ 11.26 & 78.98 $\pm$ 12.03 & 85.15 $\pm$ 12.82 \\
 & FIL & 77.84 $\pm$ 14.84 & 78.05 $\pm$ 14.96 & 76.69 $\pm$ 16.41 \\
\bottomrule
\end{tabular}
\end{table}

\FloatBarrier
\end{document}